\documentclass[MGempty]{jmlr_MG}

\usepackage{longtable}

\usepackage{booktabs}

\newcommand{\vc}[1]{\boldsymbol{\mathbf{#1}}}
\DeclareMathOperator*{\argmin}{argmin}
\DeclareMathOperator*{\oalpha}{\overline{\alpha}}
\definecolor{light-gray}{gray}{0.7}
\newcommand{\lightbox}{\color{light-gray}\rule{0.23cm}{0.09cm}}
\usepackage{algorithmic}
\usepackage{wrapfig}
\usepackage{paralist}


\jmlrworkshop{Accepted to ACML 2017}

\title[Learning Predictive Leading Indicators]{Learning Predictive Leading Indicators for Forecasting Time Series Systems with Unknown Clusters of Forecast Tasks}

 \author{\Name{Magda Gregorov\'a}$^{1,2}$ \Email{magda.gregorova@hesge.ch}\\ 
 \Name{Alexandros Kalousis}$^{1,2}$ \Email{alexandros.kalousis@hesge.ch}\\
 \Name{St\'ephane Marchand-Maillet}$^{2}$ \Email{stephane.marchand-maillet@unige.ch} \\
   \addr $^1$Geneva School of Business Administration, HES-SO University of Applied Sciences of Western Switzerland;
    $^2$University of Geneva, Switzerland}

\begin{document}

\maketitle

\begin{abstract}
We present a new method for forecasting systems of multiple interrelated time series.
The method learns the forecast models together with discovering leading indicators from within the system that serve as good predictors improving the forecast accuracy and a cluster structure of the predictive tasks around these.
The method is based on the classical linear vector autoregressive model (VAR) and links the discovery of the leading indicators to inferring sparse graphs of Granger causality.
We formulate a new constrained optimisation problem to promote the desired sparse structures across the models and the sharing of information amongst the learning tasks in a multi-task manner.
We propose an algorithm for solving the problem and document on a battery of synthetic and real-data experiments the advantages of our new method over baseline VAR models as well as the state-of-the-art sparse VAR learning methods.
\end{abstract}

\section{Introduction}
\label{sec:intro}

Time series forecasting is vital in a multitude of application areas.
With the increasing ability to collect huge amounts of data, users nowadays call for forecasts for large systems of series.
On one hand, practitioners typically strive to gather and include into their models as many potentially helpful data as possible.
On the other hand, the specific domain knowledge rarely provides sufficient understanding as to the relationships amongst the series and their importance for forecasting the system.
This may lead to cluttering the forecast models with irrelevant data of little predictive benefit thus increasing the complexity of the models with possibly detrimental effects on the forecast accuracy (over-parametrisation and over-fitting).

In this paper we focus on the problem of forecasting such large time series systems from their past evolution.
We develop a new forecasting method that learns sparse structured models taking into account the unknown underlying relationships amongst the series.
More specifically, the learned models use a limited set of series that the method identifies as useful for improving the predictive performance.
We call such series the \emph{leading indicators}.

In reality, there may be external factors from outside the system influencing the system developments.
In this work we abstract from such external con-founders for two reasons.
First, we assume that any piece of information that could be gathered has been gathered and therefore even if an external confounder exists, there is no way we can get any data on it.
Second, some of the series in the system may serve as surrogates for such unavailable data and we prefer to use these to the extent possible rather than chase the holy grail of full information availability.

We focus on the class of linear vector autoregressive models (VARs) which are simple yet theoretically well-supported, 
and well-established in the forecasting practice as well as the state-of-the-art time series literature, e.g. \cite{Lutkepohl2005}.
The new method we develop falls into the broad category of graphical-Granger methods, e.g. \cite{Lozano2009,Shojaie2010,Songsiri2013}.
Granger causality \citep{Granger1969} is a notion used for describing a specific type of dynamic dependency between time series.
In brief, a series $Z$ Granger-causes series $Y$ if, given all the other relevant information, we can predict $Y$ more accurately when we use the history of $Z$ as an input in our forecast function. 
In our case, we call such series $Z$, that contributes to improving the forecast accuracy, the leading indicator.

For our method, we assume little to no prior knowledge about the structure of the time series systems.
%
Yet, we do assume that most of the series in the system bring, in fact, no predictive benefit for the system, and that there are only few leading indicators whose inclusion into the forecast model as inputs improves the accuracy of the forecasts.
Technically this assumption of only few leading indicators translates into a sparsity assumption for the forecast model,
more precisely, sparsity in the connectivity of the associated Granger-causal graph.

An important subtlety for the model assumptions is that the leading indicators may not be \emph{leading} for the whole system but only for some parts of it (certainly more realistic especially for lager systems).
A series $Z$ may not Granger-cause all the other series in the system but only some of them.
Nevertheless, if it contributes to improving the forecast accuracy of a group of series, we still consider it a leading indicator for this group.
In this sense, we assume the system to be composed of clusters of series organised around their leading indicators.
However, neither the identity of the leading indicators nor the composition of the clusters is known a priori.

To develop our method, we built on the paradigms of multi-task, e.g. \cite{Caruana1997,Evgeniou2004}, and sparse structured learning \citep{Bach2012}.
In order to achieve higher forecast accuracy our method encourages the tasks to borrow strength from one another during the model learning.
More specifically, it intertwines the individual predictive tasks 
by shared structural constraints derived from the assumptions above.

To the best of our knowledge this is the first VAR learning method that promotes common sparse structures across the forecasting tasks of the time series system in order to improve the overall predictive performance.
We designed a novel type of structured sparsity constraints coherent with the structural assumptions for the system, integrated them into a new formulation of a VAR optimisation problem, and proposed an efficient algorithm for solving it.
The new formulation is unique in being able to discover clusters of series based on the structure of their predictive models concentrated around small number of leading indicators.



\paragraph{Organisation of the paper} The following section introduces more formally the basic concepts: linear VAR model and Granger causality.
The new method is described in section \ref{sec:newmethods}.
For clarity of exposition we start in section \ref{sec:scvar} from a set of simplified assumptions. 
The full method for learning VAR models with task Clustering around Leading indicators (CLVAR) is presented in section \ref{sec:CLVAR}.
We review the related work in section \ref{sec:relatedwork}.
In section \ref{sec:experiments} we present the results of a battery of synthetic and real-data experiments in which we confirm the good performance of our method as compared to a set of baseline state-of-the-art methods.
We also comment on unfavourable configurations of data and the bottlenecks in scaling properties.
We conclude in section \ref{sec:conclusions}.


\hspace{5em}

\section{Preliminaries}
\label{sec:preliminaries}

\paragraph{Notation}
We use bold upper case and lower case letters for matrices and vectors respectively, and plain letters for scalars (including elements of vectors and matrices).
For a matrix $\vc{A}$, the vectors $\vc{a}_{i,.}$ and $\vc{a}_{.,j}$ indicate its $i$th row and $j$th column, $a_{i,j}$ is the $(i,j)$ element of the matrix. 
$\vc{A}'$ is the transpose of $\vc{A}$, $diag(\vc{A})$ is the matrix constructed from the diagonal of $\vc{A}$, $\odot$ is the Hadamard product, $\otimes$ is the Kronecker product, $vec(\vc{A})$ is the vectorization operator, and $||\vc{A}||_F$ is the Frobenius norm.
Vectors are by convention column-wise so that $\vc{x} = (x_1, \ldots, x_n)'$ is the $n$-dimensional vector $\vc{x}$.
For any vectors $\vc{x, y}$, $\langle \vc{x},\vc{y} \rangle$ and $||\vc{x}||_2$ are the standard inner product and $\ell_2$ norms.
$\vc{1}_K$ is the $K$-dimensional vector of ones.




\subsection{Vector Autoregressive Model}
\label{sec:var}

For a set of $K$ time series observed at $T$ synchronous equidistant time points we write the VAR in the form of a multi-output regression problem as $\vc{Y} = \vc{X W} + \vc{E}$.
Here $\vc{Y}$ is the $T \times K$ output matrix for $T$ observations and $K$ time series as individual 1-step-ahead forecasting tasks, $\vc{X}$ is the $T \times Kp$ input matrix so that each row $\vc{x}_{t,.}$ is a $Kp$ long vector with $p$ lagged values of the K time series as inputs 
 $\vc{x}_{t,.} = (y_{t-1,1},y_{t-2,1},\ldots,y_{t-p,1},y_{t-1,2},\ldots,y_{t-p,K})'$,
and $\vc{W}$ is the corresponding $Kp \times K$ parameters matrix where each column $\vc{w}_{.,k}$ is a model for a single time series forecasting task (see Fig. \ref{fig:WMatrix}).
We follow the standard time series assumptions: the $T \times K$ error matrix $\vc{E}$ is a random noise matrix with i.i.d. rows with zero mean and a diagonal covariance; the time series are second order stationary and centred (so that we can omit the intercept).

In principle, we can estimate the model parameters by minimising the standard squared error loss
\begin{equation}\label{eq:SquarredLoss}
\mathit{L}(\vc{W}):= \sum_{t=1}^T \sum_{k=1}^K (y_{t,k} - \langle \vc{w}_{.,k},\vc{x}_{t,.} \rangle )^2
\end{equation} 
which corresponds to maximising the likelihood with i.i.d. Gaussian errors and spherical covariance.
However, since the dimensionality $Kp$ of the regression problem quickly grows with the number of series $K$ (by a multiple of $p$), often even relatively small VARs suffer from over-parametrisation ($Kp \gg T$).
Yet, typically not all the past of all the series is indicative of the future developments of the whole system.
In this respect the VARs are typically sparse.

In practice, the univariate autoregressive model (AR) which uses as input for each time series forecast model only its own history (and thus is an extreme sparse version of VAR), is often difficult to beat by any VAR model with the complete input sets.
A variety of approaches such as Bayesian or regularisation techniques have been successfully used in the past to promote sparsity and condition the model learning.
Those most relevant to our work are discussed in section \ref{sec:relatedwork}.



\subsection{Granger-causality Graphs}
\label{sec:granger}

\cite{Granger1969} proposed a practical definition of causality in time series based on the accuracy of least-squares predictor functions.
In brief, for two time series $Z$ and $Y$, we say that $Z$ Granger causes if, given all the other relevant information, a predictor function using the history of $Z$ as input can forecast $Y$ better (in the mean-square sense) than a function not using it. 
Similarly, a set of time series $\{Z_1, \ldots, Z_l\}$ G-causes series $Y$ if it can be predicted better using the past values of the set.

The G-causal relationships can be described by a directed graph $\mathcal{G} = \{\vc{\mathcal{V}},\vc{\mathcal{E}}\}$ (\cite{Eichler2012}), where 
each node $v \in \vc{\mathcal{V}}$ represents a time series in the system, 
and the directed edges represent the G-causal relationships between the series.
In VARs the G-causality is captured within the $\vc{W}$ parameters matrix. 
When any of the parameters of the $k$-th task ($k$-th column of the $\vc{W}$) referring to the $p$ past values of the $l$-th input series is non-zero, we say that the $l$-th series G-causes series $k$, and we denote this in the G-causal graph by a directed edge $e_{l,k}$ from $v_l$ to $v_k$. 




\begin{wrapfigure}{r}{0.45\textwidth}
\vspace{-25pt}
  \begin{center}
    \input{WMatrix1}
  \end{center}
\vspace{-10pt}
\caption{$\vc{W}$ and G-causal graph.}\label{fig:WMatrix}
\end{wrapfigure}

Fig. \ref{fig:WMatrix} shows a schema of the VAR parameters matrix $\vc{W}$ and the corresponding G-causal graph for an example system of $K=7$ series with the number of lags $p=3$. 
In \ref{fig:WMatrix}(a) the gray cells are the non-zero elements, in \ref{fig:WMatrix}(b) the circle nodes are the individual time series, the arrow edges are the G-causal links between the series\footnote{The self-loops corresponding to the block-diagonal elements in $\vc{W}$ are omitted for clarity of display.}.
For example, the arrow from 2 to 1 indicates that series 2 G-causes series 1;
correspondingly the cells for the 3 lags in the 2nd block-row and the 1th column are shaded ($\vc{\widetilde{w}}_{2,1}$).  Series 2 and 5 are the leading indicators for the whole system, their block-rows are shaded in all columns in the $\vc{W}$ matrix schema and they have out-edges to all other nodes in the G-graph. 

One may question if calling the above notion \emph{causality} is appropriate. 
Indeed, unlike other perhaps more philosophical approaches, e.g. \cite{Pearl2009}, it does not really seek to understand the underlying forces driving the relationships between the series. Instead, the concept is purely technical based on the series contribution to the predictive accuracy, ignoring also possible confounding effects of unobservables.
Nevertheless, the term is well established in the time series community. 
Moreover, it fits very well our purposes, where the primary objective is to learn models with high forecast accuracy that use as inputs only those time series that contribute to improving the accuracy - the leading indicators.
Therefore, acknowledging all the reservations, we stick to it in this paper always preceding it by Granger or G- to avoid confusion.  

\section{Learning VARs with Clusters around Leading Indicators}
\label{sec:newmethods}

We present here our new method for learning VAR models with task Clustering around Leading indicators (CLVAR).
The method relies on the assumption that the generating process is sparse in the sense of there being only a few leading indicators within the system having an impact on the future developments.
The leading indicators may be useful for predicting all or only some of the series in the systems.
In this respect the series are clustered around their G-causing leading indicators.
However, the method does not need to know the identity of the leading indicators nor the cluster assignments a priori and instead learns these together with the predictive models.



In building our method we exploited the multi-task learning ideas \citep{Caruana1997} and let the models benefit from learning multiple tasks together (one task per series).
This is in stark contrast to other state-of-the-art VAR and graphical-Granger methods, e.g. \cite{Arnold2007,Lozano2009,Liu2012}.
Albeit them being initially posed as multi-task (or multi-output) problems, due to their simple additive structure they decompose into a set of single-task problems solvable independently without any interaction and information sharing during the per-task learning.
We, on the other hand, encourage the models to share information and borrow strength from one another in order to improve the overall performance by intertwining the model learning via structural constraints on the models derived from the assumptions outlined above.
%


\subsection{Leading Indicators for Whole System}
\label{sec:scvar}

For the sake of exposition we first concentrate on a simplified problem of learning a VAR with leading indicators shared by the whole system (without clustering).
The structure we assume here is the one illustrated in Fig. \ref{fig:WMatrix}.
We see that the parameters matrix $\vc{W}$ is sparse with non-zero elements only in the block-rows corresponding to the lags of the leading indicators for the system (series 2 and 5 in the example in Fig. \ref{fig:WMatrix}) and on the block diagonal.
The block-diagonal elements of $\vc{W}$ are associated with the lags of each series serving as inputs for predicting its own 1-step-ahead future.
It is a stylised fact that the future of a stationary time series depends first and foremost on its own past developments. Therefore in addition to the leading indicators we want each of the individual series forecast function to use its own past as a relevant input.
We bring the above structural assumptions into the method by formulating novel fit-for-purpose constraints for learning VAR models with multi-task structured sparsity.

\subsubsection{Learning Problem and Algorithm for Learning without Clusters}
\label{sec:scvarlearningproblem}


We first introduce some new notation to accommodate for the necessary block structure across the lags of the input series in the input matrix $\vc{X}$ and the corresponding elements of the parameters matrix $\vc{W}$.
%
For each input vector $\vc{x}_{t,.}$ (a row of $\vc{X}$) we indicate by $\vc{\widetilde{x}}_{t,j} = (y_{t-1,j},y_{t-2,j},\ldots,y_{t-p,j})'$ the $p$-long sub-vector of $\vc{x}_{t,.}$ referring to the history (the $p$ lagged values preceding time $t$) of the series $j$, so that for the whole row we have  $\vc{x}_{t,.} = (x_{t,1}, \ldots, x_{t,Kp})' = (\vc{\widetilde{x}}'_{t,1}, \ldots, \vc{\widetilde{x}}'_{t,K})'$. 
Correspondingly, in each model vector $\vc{w}_{.,k}$ (a column of $\vc{W}$), we indicate by $\vc{\widetilde{w}}_{j,k}$ the $p$-long sub-vector of the $k$th model parameters associated with the input sub-vector $\vc{\widetilde{x}}_{t,j}$.
In Fig. \ref{fig:WMatrix}, $\vc{\widetilde{w}}_{2,1}$ is the block of the 3 shaded parameters in column 1 and rows $\{4,5,6\}$ - the block of parameters of the model for forecasting the 1st time series associated with the 3 lags of the 2nd time series (a leading indicator) as inputs.
Using these blocks of inputs $\vc{\widetilde{x}}_{t,j}$ and parameters $\vc{\widetilde{w}}_{j,k}$ we can rewrite the inner products in the loss in \eqref{eq:SquarredLoss} as $\langle \vc{w}_{.,k},\vc{x}_{t,.} \rangle = \sum_{b=1}^K \langle \vc{\widetilde{w}}_{b,k},\vc{\widetilde{x}}_{t,b} \rangle$.

%

Next, we associate each of the parameter blocks with a single non-negative scalar $\gamma_{b,k}$ so that $\vc{\widetilde{w}}_{b,k} = \gamma_{b,k} \, \vc{\widetilde{v}}_{b,k}$.
The $Kp \times K$ matrix $\vc{V}$, composed of the blocks $\vc{\widetilde{v}}_{b,k}$ in the same way as $\vc{W}$ is composed of $\vc{\widetilde{w}}_{b,k}$, is therefore just a rescaling of the original $\vc{W}$ with the weights $\gamma_{b,k}$ used for each block.
With this new re-parametrization the squared-error loss \eqref{eq:SquarredLoss} is
\begin{equation}\label{eq:SquarredLoss3}
\mathit{L}(\vc{W}) = \sum_{t=1}^T \sum_{k=1}^K (y_{t,k} - \sum_{b=1}^K \gamma_{b,k} \langle \vc{\widetilde{v}}_{b,k},\vc{\widetilde{x}}_{t,b} \rangle )^2.
\end{equation} 

Finally, we use the non-negative $K \times K$ weight matrix $\vc{\Gamma} = \{ \gamma_{b,k} \ | \, b,k=1,\dots,K \}$ to formulate our multi-task structured sparsity constraints.
In $\vc{\Gamma}$ each element corresponds to a single series serving as an input to a single predictive model.
A zero weight $\gamma_{b,k} = 0$ results in a zero parameter sub-vector $\vc{\widetilde{w}}_{b,k} = \vc{0}$ and therefore the corresponding input sub-vectors $\vc{\widetilde{x}}_{t,b}$ (the past lags of series $b$ for each time point $t$) have no effect in the predictive functions for task $k$.

Our assumption of only small number of leading indicators means that most series shall have no predictive effect for any of the tasks.
This can be achieved by $\vc{\Gamma}$ having most of its rows equal to zero.
On the other hand, the non-zero elements corresponding to the leading indicators shall form full rows of $\vc{\Gamma}$.
As explained in section \ref{sec:scvar}, in addition to the leading indicators we also want each series past to serve as an input to its own forecast function. 
This translates to non-zero diagonal elements $\gamma_{i,i} \neq 0$.
To combine these two contradicting structural requirements onto $\vc{\Gamma}$ (sparse rows vs. non-zero diagonal) we construct the matrix from two same size matrices $\vc{\Gamma} = \vc{A} + \vc{B}$, one for each of the structures: $\vc{A}$ for the row-sparse of leading indicators, $\vc{B}$ for the diagonal of the own history.


We now formulate the optimisation problem for learning VAR with shared leading indicators across the whole system and dependency on own past as the constrained minimisation
\begin{eqnarray}\label{eq:Optimisation1}
& \argmin _{\vc{A,V}} \ \sum_{t=1}^T \sum_{k=1}^K (y_{t,k} - \sum_{b=1}^K (\alpha_{b,k} + \beta_{b,k}) \langle \vc{\widetilde{v}}_{b,k},\vc{\widetilde{x}}_{t,j} \rangle )^2 + \lambda ||\vc{V}||_F^2 & \\
& \text{ s.t. }  \quad \vc{1}'_K \, \vc{\oalpha} = \kappa;
\, \vc{\oalpha} \geq \vc{0};
\, \vc{\alpha}_{.,j} = \vc{\oalpha}, \, \beta_{j,j} = 1 - \alpha_{j,j}  \ \forall j = 1,\ldots, K \, & \nonumber \enspace , 
\end{eqnarray}
where the links between the matrices $\vc{A,B,\Gamma,V}$ and the parameter matrix $\vc{W}$ of the VAR model are explained in the paragraphs above.

In (\ref{eq:Optimisation1}) we force all the columns of $\vc{A}$ to be equal to the same vector $\vc{\oalpha}$\footnote{This does not excessively limit the capacity of the models as the final model matrix $\vc{W}$ is the result of combining $\vc{\Gamma}$ with the learned matrix $\vc{V}$.}, and we promote the sparsity in this vector by constraining it onto a simplex of size $\kappa$. $\kappa$ controls the relative weight of each series own past vs. the past of all the neighbouring series.
For identifiability reasons we force the diagonal elements of $\vc{\Gamma}$ to equal unity by scaling appropriately the diagonal $\beta_{j,j}$ elements.
Lastly, 
while $\vc{\Gamma}$ is constructed and constrained to control for the structure of the learned models (as per our assumptions), the actual value of the final parameters $\vc{W}$ is the result of combining it with the other learned matrix $\vc{V}$. 
To confine the overall complexity of the final model $\vc{W}$ we impose a standard ridge penalty \citep{Hoerl1970} on the model parameters $\vc{V}$.

The optimisation problem \eqref{eq:Optimisation1} is jointly non-convex, however, it is convex with respect to each of the optimisation variables with the other variable fixed.
Therefore we propose to solve it by an alternating descent for $\vc{A}$ and $\vc{V}$ as outlined in algorithm \ref{alg:SharedLead} below. $\vc{B}$ is solved trivially applying directly the equality constraint of \eqref{eq:Optimisation1} over the learned matrix $\vc{A}$ as $\vc{B} = \vc{I} - diag(\vc{A})$ which implies $\vc{\Gamma} = \vc{A} + \vc{B} = \vc{A} - diag(\vc{A}) + \vc{I}$.

\begin{algorithm2e}
\caption{Alternating descent for VAR with system-shared leading indicators}\label{alg:SharedLead}
\SetKwInOut{Input}{Input}
\SetKwInOut{Initialise}{Initialise}
\SetKwBlock{Step1}{Step1}{end}
\SetKwBlock{S2}{Step2}{end}
\Input{training data $\vc{Y}, \vc{X}$; hyper-parameters $\lambda, \kappa$}
\Initialise{$\vc{\oalpha}$ evenly to satisfy constraints in all columns of $\vc{A}$;
$\vc{\Gamma} \leftarrow \vc{A} - diag(\vc{A}) + \vc{I}$
}
\BlankLine
\Repeat(\tcp*[f]{Alternating descent}) {objective convergence}{
\Begin(Step 1: Solve for $\vc{V}$){
\ForEach{task $k$}{
re-weight input blocks $\vc{z}_{t,b}^{(k)} \leftarrow  \gamma_{b,k} \, \vc{\widetilde{x}}_{t,b}$ \quad $\forall$ time point $t$ and input series $b$ \\
$\vc{v}_{.,k} \leftarrow \argmin_{\vc{v}} ||\vc{y}_{.,k}-\vc{Z}^{(k)} \, \vc{v}||_2^2 + \lambda ||\vc{v}||_2^2$
\tcp*[f]{standard ridge regression}
}
}
\Begin(Step 2: Solve for $\vc{A}$ and $\vc{\Gamma}$){
\ForEach{task $k$}{
input products $h_{t,b}^{(k)} \leftarrow \langle \vc{\widetilde{v}}_{b,k}, \vc{\widetilde{x}}_{t,b} \rangle$ \quad $\forall$ time point $t$ and input series $b$ \\
task residuals after using own history $r_{t,k} \leftarrow y_{t,k} - h_{t,k}^{(k)}$ \quad $\forall$ time point $t$ \\
remove own history from input products $h_{t,k}^{(k)} \leftarrow 0$ \quad $\forall$ time point $t$\\
}
concatenate vertically input product matrices $\vc{H} = vertcat(\vc{H}^{(.)})$
$\vc{\oalpha} \leftarrow \argmin_{\vc{\oalpha}} ||vec(\vc{R})-\vc{H}\,\vc{\oalpha}||_2^2$, s.t. $\vc{\oalpha}$ on simplex 
\tcp*[f]{projected grad descent}
put $\vc{\oalpha}$ to all columns of $\vc{A}$;
$\vc{\Gamma} \leftarrow \vc{A} - diag(\vc{A}) + \vc{I}$
}
}
\end{algorithm2e}


To foster the intuition behind our method we provide links to other well-known learning problems and methods.
First, we can rewrite the weighted inner product in the loss function (\ref{eq:SquarredLoss3}) as $\langle \vc{\widetilde{v}}_{b,k},\gamma_{b,k} \vc{\widetilde{x}}_{t,b} \rangle$. In this ``feature learning'' formulation the weights $\gamma_{b,k}$ act on the original inputs and, hence, generate new task-specific features $\vc{z}_{t,b}^{(k)} =  \gamma_{b,k} \, \vc{\widetilde{x}}_{t,b}$. These are actually used in Step 1 of our algorithm \ref{alg:SharedLead}.
Alternatively, we can express the ridge penalty on $\vc{V}$ used in eq. ({\ref{eq:Optimisation1}}) as
$||\vc{V}||_F^2 = \sum_{b,k} ||\vc{\widetilde{v}}_{b,k}||_2^2 = \sum_{b,k} 1/\gamma_{b,k}^2 ||\vc{\widetilde{w}}_{b,k}||_2^2$. In this ``adaptive ridge'' formulation the elements of $\vc{\Gamma}$, which in our methods we learn, act as weights for the $\ell_2$ regularization of $\vc{W}$.
Equivalently, we can see this as the Bayesian maximum-a-posteriori with Guassian priors where the elements of $\vc{\Gamma}$ are the learned priors for the variance of the model parameters or (perhaps more interestingly) the random errors.


\subsection{Leading Indicators for Clusters of Predictive Tasks}
\label{sec:CLVAR}

After explaining in section \ref{sec:scvar} the simplified case of learning a VAR with leading indicators for the whole system, we now move onto the more complex (and for larger VARs certainly more realistic) setting of the leading indicators being predictive only for parts of the system - clusters of predictive tasks.


To get started we briefly consider the situation in which the cluster structure (not the leading indicators) is known a priori.
Here the models could be learned by a simple modification of algorithm \ref{alg:SharedLead} where in step 2 we would work with cluster-specific vectors $\vc{\oalpha}$ and matrices $\vc{H}$ and $\vc{R}$ constructed over the known cluster members.
%
%
In reality the clusters are typically not known and therefore our CLVAR method is designed to learn them together with the leading indicators.

We use the same block decompositions of the input and parameter matrices $\vc{X}$ and $\vc{W}$, and the structural matrices $\vc{\Gamma} = \vc{A} + \vc{B} = \vc{A} - diag(\vc{A}) + \vc{I}$ and the rescaled parameter matrix $\vc{V}$ defined in section \ref{sec:scvar}. 
However, we need to alter the structural assumptions encoded into the matrix $\vc{A}$.
In the cluster case $\vc{A}$ still shall have many rows equal to zero but it shall no longer have all the columns equal (same leading indicators for all the tasks). 
Instead, we learn it as a low rank matrix by factorizing it into two lower dimensional matrices $\vc{A} = \vc{DG}$: the $K \times r$ dictionary matrix $\vc{D}$ with the dictionary atoms (columns of $\vc{D}$) representing the cluster prototypes of the dependency structure; and the $r \times K$ matrix $\vc{G}$ with the elements being the per-model dictionary weights, $1 \leq r \leq K$.

\begin{wrapfigure}{r}{0.5\textwidth}
\vspace{-10pt}
\centering
\subfigure[hard cluster assignments][b]{\label{fig:DGsub1}\includegraphics[width=0.2\textwidth]{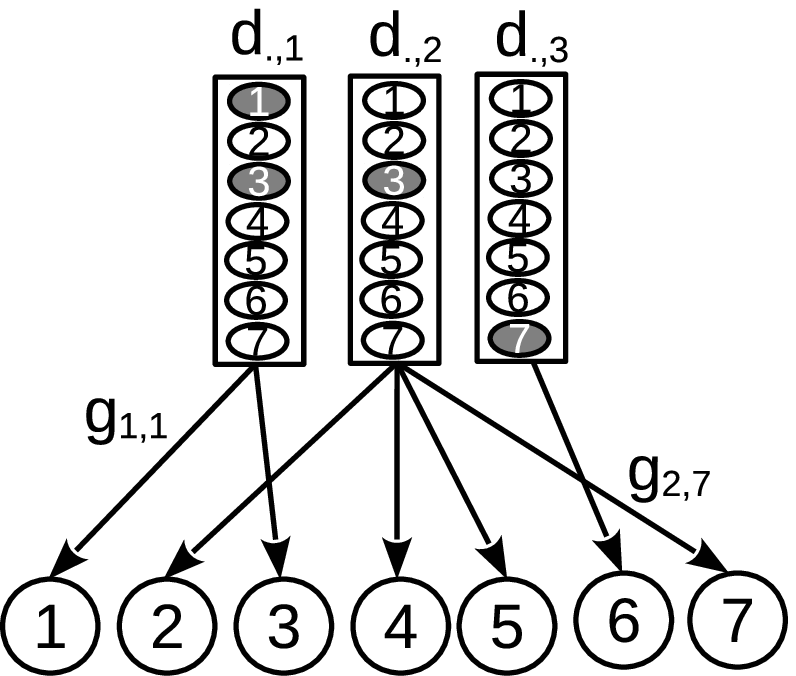}}
\quad
\subfigure[soft cluster assignments][b]{\label{fig:DGsub2}\includegraphics[width=0.2\textwidth]{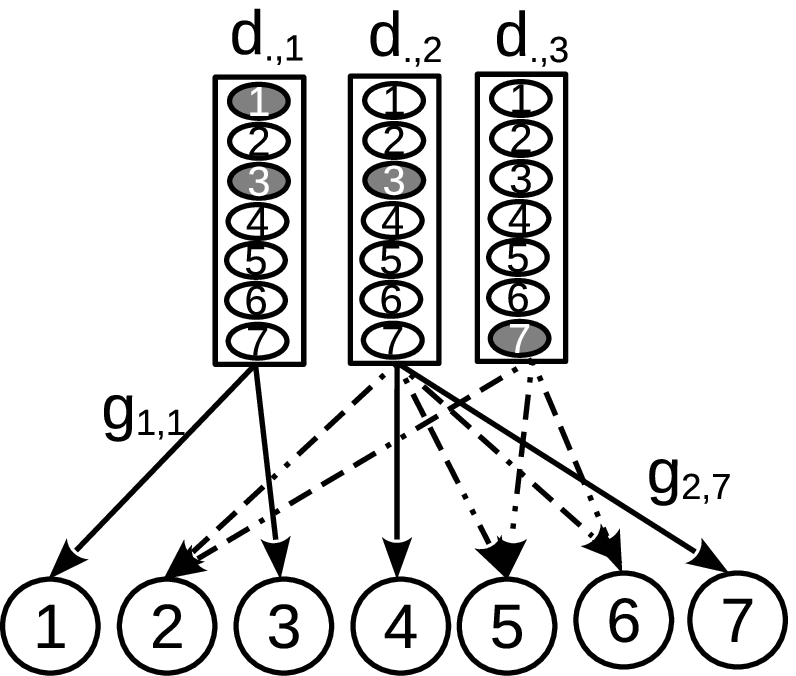}}
\caption{Roles of $\vc{D}$ and $\vc{G}$ matrices in the low-rank decomposition $\vc{A}$}\label{fig:DG}
\vspace{-10pt}
\end{wrapfigure}

To better understand the clustering effect of the low-rank decomposition, Fig. \ref{fig:DG} illustrates it for an imaginary system of $K=7$ time series with rank $r=3$.
The $\vc{d}_{.,j}$ j=\{1,2,3\} columns in the top are the sparse cluster prototypes (the non-zero elements for the leading indicators are shaded).
The circles in the bottom are the individual learning tasks and the arrows are the per-model dictionary weights $g_{i,j}$.
Solid arrows have weight 1, missing arrows have weight zero, dashed arrows have weight between 0 and 1.
So for example, the solid arrow from the 2nd column to the 7th circle in Fig. \ref{fig:DGsub1} is the $g_{2,7}$ element of matrix $\vc{G}$. 
Since it is a full arrow, it is equal to 1.
The arrow from the 3rd column to the 2nd circle in Fig. \ref{fig:DGsub2} is the $g_{3,2}$ element of $\vc{G}$. Since the arrow is dashed, we have $0 < g_{3,2} < 1$.

Fig. \ref{fig:DGsub1} uses a binary matrix $\vc{G}$ (no dashed arrows) reflecting hard clustering of the tasks consistent with our initial setting of a priori known clusters.
Each task (circle at the bottom) is associated with only one cluster prototype (columns of $\vc{D}$ in the top). 
In contrast, Fig. \ref{fig:DGsub2} uses matrix $\vc{G}$ with elements between 0 and 1 to perform soft clustering of the tasks.
Each task (circle at the bottom) may be associated with more than one cluster prototype (columns of $\vc{D}$ in the top). Our CLVAR is based on this latter approach of soft-clustering of the forecast tasks.


\subsubsection{Learning Problem and Algorithm for CLVAR}
\label{sec:CLVARlearningproblem}

We now adapt the minimisation problem \eqref{eq:Optimisation1} for the multi-cluster setting
\begin{eqnarray}\label{eq:Optimisation2}
& \argmin _{\vc{D,G,V}} \ \sum_{t=1}^T \sum_{k=1}^K (y_{t,k} - \sum_{b=1}^K (\sum_{j=1}^K d_{b,j}g_{j,k} + \beta_{b,k}) \langle \vc{\widetilde{v}}'_{b,k},\vc{\widetilde{x}}'_{t,j} \rangle )^2  + \lambda ||\vc{V}||_F^2 & \\
& \text{ s.t. } \vc{1}'_K \, \vc{d}_{.,j} = \kappa ; \; \vc{d}_{.,j} \geq \vc{0}  ;
\; \vc{1}'_r \, \vc{g}_{.,j} = 1  ; \; \vc{g}_{.,j} \geq \vc{0}, \, \; \beta_{j,j} = 1 - \alpha_{j,j}  \ \forall j \enspace .& \nonumber
\end{eqnarray}
The relations of the optimisation matrices $\vc{D,G,V}$ to the parameter matrix $\vc{W}$ of the VAR model are as explained in the paragraphs above.
The principal difference of the formulation \eqref{eq:Optimisation2} as compared to problem \eqref{eq:Optimisation1} is the low-rank decomposition of matrix $\vc{A} = \vc{DG}$ using the fact that $a_{b,k} = \sum_{j=1}^K d_{b,j}g_{j,k}$.
Similarly as for the single column $\vc{\oalpha}$ in \eqref{eq:Optimisation1} we promote sparsity in the cluster prototypes $\vc{d}_{.,j}$ by constraining them onto the simplex. 
And we use the probability simplex constraints to sparsify the per-task weights in the columns of $\vc{G}$ so that the task are not based on all the prototypes.

\begin{algorithm2e}
\caption{CLVAR - VAR with leading indicators for clusters of predictive tasks}\label{alg:CLVAR}
\SetKwInOut{Input}{Input}
\SetKwInOut{Initialise}{Initialise}
\SetKwBlock{Step1}{Step1}{end}
\SetKwBlock{S2}{Step2}{end}
\Input{training data $\vc{Y}, \vc{X}$; hyper-parameters $\lambda, \kappa, r$}
\Initialise{$\vc{D,G}$ evenly to satisfy the constraints;
$\vc{A} \leftarrow \vc{DG}$;
$\vc{\Gamma} \leftarrow \vc{A} - diag(\vc{A}) + \vc{I}$
}
\BlankLine
\Repeat(\tcp*[f]{Alternating descent}) {objective convergence}{
\Begin(Step 1: Solve for $\vc{V}$){
same as in algorithm \ref{alg:SharedLead}
}
\Begin(Step 2: Solve for $\vc{D,G}$ and $\vc{\Gamma}$){
\ForEach{task $k$}{
same as in algorithm \ref{alg:SharedLead} \\
$\vc{g_{.,k}} \leftarrow \argmin_{\vc{g}} ||\vc{r}_{.,k}-\vc{H}^{(k)}\,\vc{g}||_2^2$, s.t. $\vc{g}$ on simplex \tcp*[f]{projected grad desc}
}
concatenate vertically input product matrices $\vc{H} = vertcat(\vc{H}^{(.)})$ \\
expand matrices to match dictionary vectorization $\vc{\widehat{G}} \leftarrow \vc{G}' \otimes \vc{1}_T \vc{1}'_K$; $\vc{\widehat{H}} = \vc{1}'_r \otimes \vc{H}$ \\
$vec(\vc{D}) \leftarrow \argmin_{\vc{D}} ||vec(\vc{R}) - \vc{\widehat{G}} \odot \vc{\widehat{H}} \, vec(\vc{D})||_2^2$  \tcp*[f]{projected grad desc} \\
\qquad \qquad \qquad \qquad \qquad s.t. $\vc{d}_{.,j}$ on simplex $\forall j$ \\
$\vc{A} = \vc{DG}$; $\vc{\Gamma} \leftarrow \vc{A} - diag(\vc{A}) + \vc{I}$
}
}
\end{algorithm2e}

We propose to solve problem \eqref{eq:Optimisation2} by alternating descent algorithm \ref{alg:CLVAR}.
While non-convex, the alternating approach for learning the low-rank matrix decomposition is known to perform well in practice and has been recently supported by new theoretical guarantees, e.g. \cite{Park2016}. 
We solve the two sub-problems in step 2 by projected gradient descent with FISTA backtracking line search \citep{Beck2009}.
The algorithm is $\mathcal{O}(T)$ for increasing number of observation and $\mathcal{O}(K^3)$ for increasing number of time series.
However, one needs to bear in mind that with each additional series the complexity of the VAR model itself increases by $\mathcal{O}(K)$.
Nevertheless, the expensive scaling with $K$ is an important bottleneck of our method and we are investigating options to address it in our future work.

%
%

\section{Related Work}
\label{sec:relatedwork}

We explained in section \ref{sec:granger} how our search for leading indicators links to the Granger causality discovery in VARs.
As shows the list of references in the survey of \cite{Liu2012}, this has been a rather active research area over the last several years.
While the traditional approach for G-discovery was based on pairwise testing of candidate models or the use of model selection criteria such as AIC or BIC, inefficiency of such approaches for builidng predictive models of large time series system has long been recognised\footnote{Due to the lack of domain knowledge to support the model selection and combinatorial complexity of exhaustive search.}, e.g. \cite{Doan1984}.

As an alternative, variants of so-called graphical Granger methods based on regularization for parameter shrinkage and thresholding (along the lines of Lasso \citep{Tibshirani1996}) have been proposed in the literature.
%
We use the two best-established ones, the lasso-Granger (VARL1) of \cite{Arnold2007} and the grouped-lasso-Granger (VARLG) of \cite{Lozano2009}, as the state-of-the-art competitors in our experiments.
More recent adaptations of the graphical Granger method address the specific problems of determining the order of the models and the G-causality simultaneously \citep{Shojaie2010,Ren2013}, the G-causality inference in irregular \citep{Bahadori2012} and subsampled series \citep{Gong2015}, and in systems with instantaneous effects \citep{Peters2013}.
However, neither of the above methods considers or exploits any common structures in the G-causality graphs as we do in our method.

Common structures in the dependency are assumed by \cite{Jalali2012} and \cite{Geiger2014} though the common interactions are with unobserved variables from outside the system rather then within the system itself. Also, the methods discussed in these have no clustering ability.
\cite{Songsiri2015} considers common structures across several datasets (in panel data setting) instead of within the dynamic dependencies of a single dataset.
\cite{Huang2012} assume sparse bi-clustering of the G-graph nodes (by the in- and out- edges) to learn fully connected sub-graphs in contrast to our shared sparse structures.
Most recently, \cite{Hong2017} proposes to learn clusters of series by Laplacian clustering over the sparse model parameters. However, the underlying models are treated independently not encouraging any common structures at learning.

More broadly, our work builds on the multi-task \citep{Caruana1997} and structured sparsity \citep{Bach2012} learning techniques developed outside the time-series settings.
Similar block-decompositions of the feature and parameter matrices as we use in our methods have been proposed to promote group structures across multiple models \citep{Argyriou2008,Swirszcz2012}. Although the methods developed therein have no clustering capability.
Various approaches for learning model clusters are discussed in \cite{Bakker2003,Xue2007,Jacob2008,Kang2011,Kumar2012} of which the latest uses similar low-rank decomposition approach as our method.
Nevertheless, neither of these approaches learns sparse models and builds the clustering on similar structural assumptions as our method does.

\section{Experiments}
\label{sec:experiments}

We present here the results of a set of experiments on synthetic and real-world datasets. 
We compare to relevant baseline methods for VAR learning: univariate auto-regressive model AR (though simple, AR is typically hard to beat by high-dimensional VARs when the domain knowledge cannot help to specify a relevant feature subset for the VAR model), VAR model with standard $\ell_2$ regularisation VARL2 (controls over-parametrisation by shrinkage but does not yield sparse models), VAR model with $\ell_1$ regularisation VARL1 (lasso-Granger of \cite{Arnold2007}), and VAR with group lasso regularisation VARLG (grouped-lasso-Granger of \cite{Lozano2009}).
We implemented all the methods in Matlab using standard state-of-the-art approaches: trivial analytical solutions for AR and VARL2, FISTA proximal-gradient \citep{Beck2009} for VARL1 and VARLG.
The full code together with the datasets amenable for full replication of our experiments is available from  \url{https://bitbucket.org/dmmlgeneva/var-leading-indicators}.

In all our experiments we simulated real-life forecasting exercises.
We split the analysed datasets into training and hold-out sets unseen at learning and only used for performance evaluation.
The trained models were used to produce one-step ahead forecasts by sliding through all the points in the hold-out.
We repeated each experiments over 20 random re-samples. 
The reported performance is the averages over these 20 re-samples.
The construction of the re-samples for the synthetic and real datasets is explained in the respective sections below.
We used 3-folds cross-validation with mean squared error as the criterion for the hyper-parameter grid search. Unless otherwise stated below, the grids were: $\lambda \in$ 15-elements grid $[10^{-4} \ldots 10^{3}]$ (used also for VARL2, VARL1 and VARLG), $\kappa \in \{0.5, 1, 2\}$, rank $\in \{1, 0.1K, 0.2K, K\}$.
We preprocessed all the data by zero-centering and unit-standardization based on the training statistics only.

For all the experiments and all the tested methods we fixed the lag of the learned models to $p=5$.
While the search for the best lag $p$ has in the past constituted an important part of time series modelling\footnote{Especially for univariate models within the context of the more general ARMA class \citep{Box1994}.}, in high-dimensional settings the exhaustive search through possible sub-set model combinations is clearly impractical.
Modern methods therefore focus on using VARs with sufficient number of lags to cater for the underlying time dependency and apply Bayesian or regularization methods to control the model complexity, e.g. \cite{Koop2013}.
In our case, this is achieved by the ridge shrinkage on the parameter matrix $\vc{V}$.

\subsection{Synthetic Experiments}
\label{sec:syntheticexperiments}

We designed six generating processes for systems varying by number of series and the G-causal structure.
The first three are small systems with $K=10$ series only, the next three increase the size to $K=\{30, 50, 100\}$.
Systems 1 and 2 are unfavourable for our method, generated by processes not corresponding to our structural assumptions: in the 1st each series is generated from its own past only and therefore can be best modelled by a simple univariate AR model (the G-causal graph has no links); the 2nd is a fully connected VAR (all series are leading for the whole system).
The 3rd system consists of 2 clusters with 5 series each, both depending on 1 leading indicator.
Systems 4-6 are composed of $\{3,5,10\}$ clusters respectively, each with 10 series concentrated around 2 leading indicators\footnote{For the last two, we fixed the rank in CLVAR training to the true number of clusters.}.


For each of the 6 system designs we first generated a random matrix of VAR coefficients with the required structure. We ensured the processes are stationary by controlling the roots of the model characteristic polynomials.  
We then generated 20 random realisation of the VAR processes with uncorrelated standard-normal noise.
In each, we separated the last 500 observations into a hold-out set and used the previous $T$ observations for training. 
Once trained, the same model was used for the 1-step-ahead forecasting of the 500 hold-out points by sliding forward through the dataset.

The predictive performance of the methods in the 6 experimental settings for multiple training sizes $T$ is summarised in Fig. \ref{fig:synthMse}\footnote{Numerical results behind the plots are listed in the Supplement.}.
We measure the predictive accuracy by the mean square error of 1-step-ahead forecasts relative to the forecasts produced by the VAR with the true generative coefficients (RelMSE). Doing so we standardize the MSE by the irreducible error of each of the forecast exercises. 
The closer to 1 (the gold standard) the better.
The plots display the average RelMSE over the twenty replications of the experiments, the error bars are at $\pm 1$ standard deviation.

In all the experiments the predictive performance improves with the increasing training size and the differences between the methods diminish. 
CLVAR outperforms all the other methods in the experiments with sparse structures as per our assumptions (mostly markedly).
But CLVAR behaves well even in the unfavourable conditions of the first two systems. 
It still performs better than the other two sparse methods VARL1 and VARLG and the non-sparse VARL2 in the 1st completely sparse experiment\footnote{The AR model is in an advantage here since it has the true-process structure by construction.},
and it is on par with the other methods in the 2nd full VAR experiment.

\begin{figure}[h!]
\centering
\subfigure[Relative MSE over true model][b]{\label{fig:synthMse}\includegraphics[width=.49\textwidth]{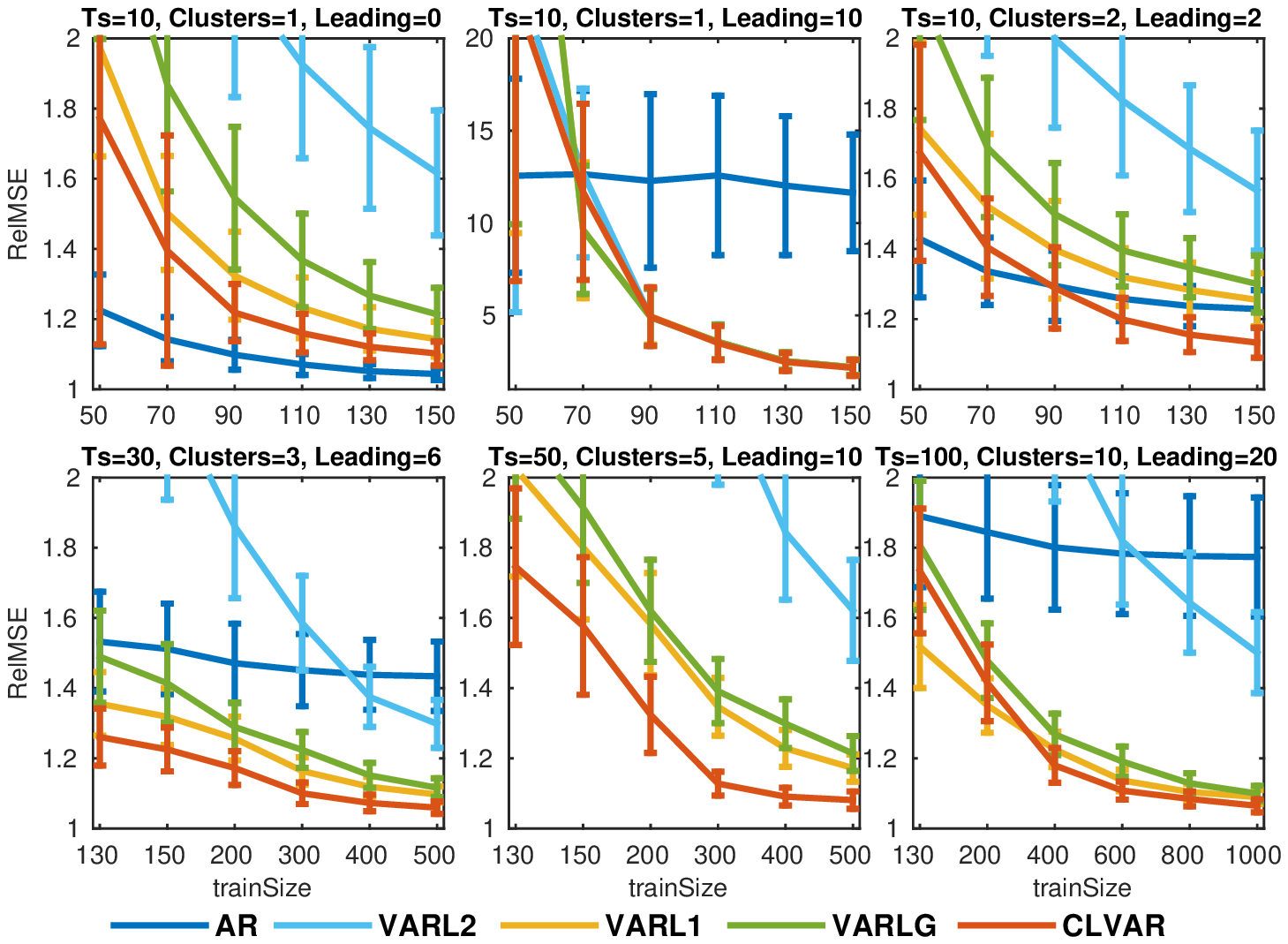}}
\subfigure[Selection error of G-causal links][b]{\label{fig:synthFnrFpr}\includegraphics[width=.49\textwidth]{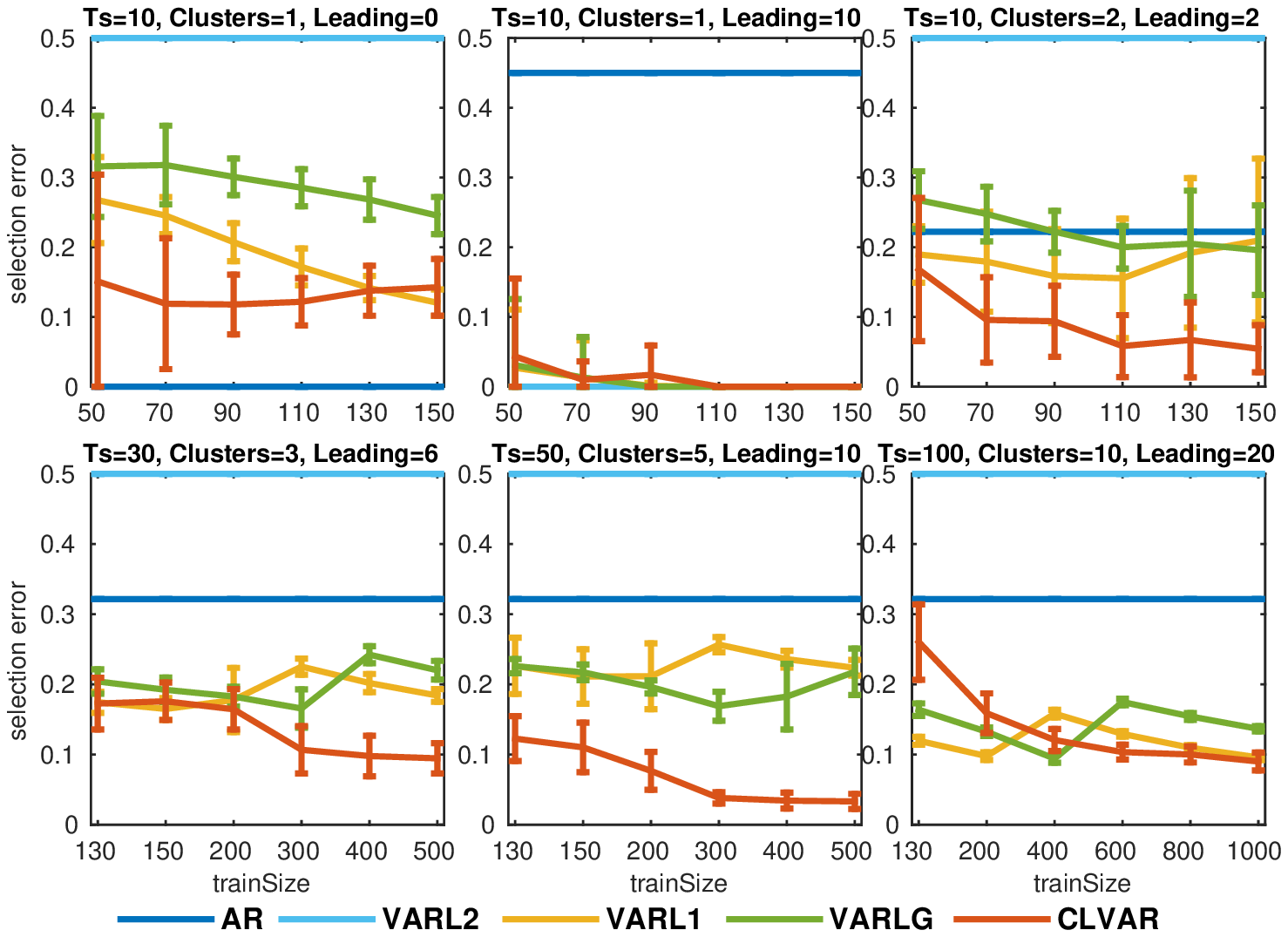}}
\caption{Results for synthetic experiments averaged over 20 experimental replications}\label{fig:Synth}
\end{figure}


In Fig. \ref{fig:synthFnrFpr} we show the accuracy of the methods in selecting the true generative G-causal links between the series in the system.
The selection error  (the lower the better) is measured as the average of the false negative and false positive rates.
We plot the averages with $\pm 1$ standard deviation over the 20 experimental replications.
The CLVAR typically learned models structurally closer to the true generating process than the other tested methods, in most cases with substantial advantage.


\begin{wrapfigure}{r}{0.5\textwidth}
\vspace{-10pt}
\centering{\includegraphics[trim={370 132 0 140},clip,width=0.4\textwidth]{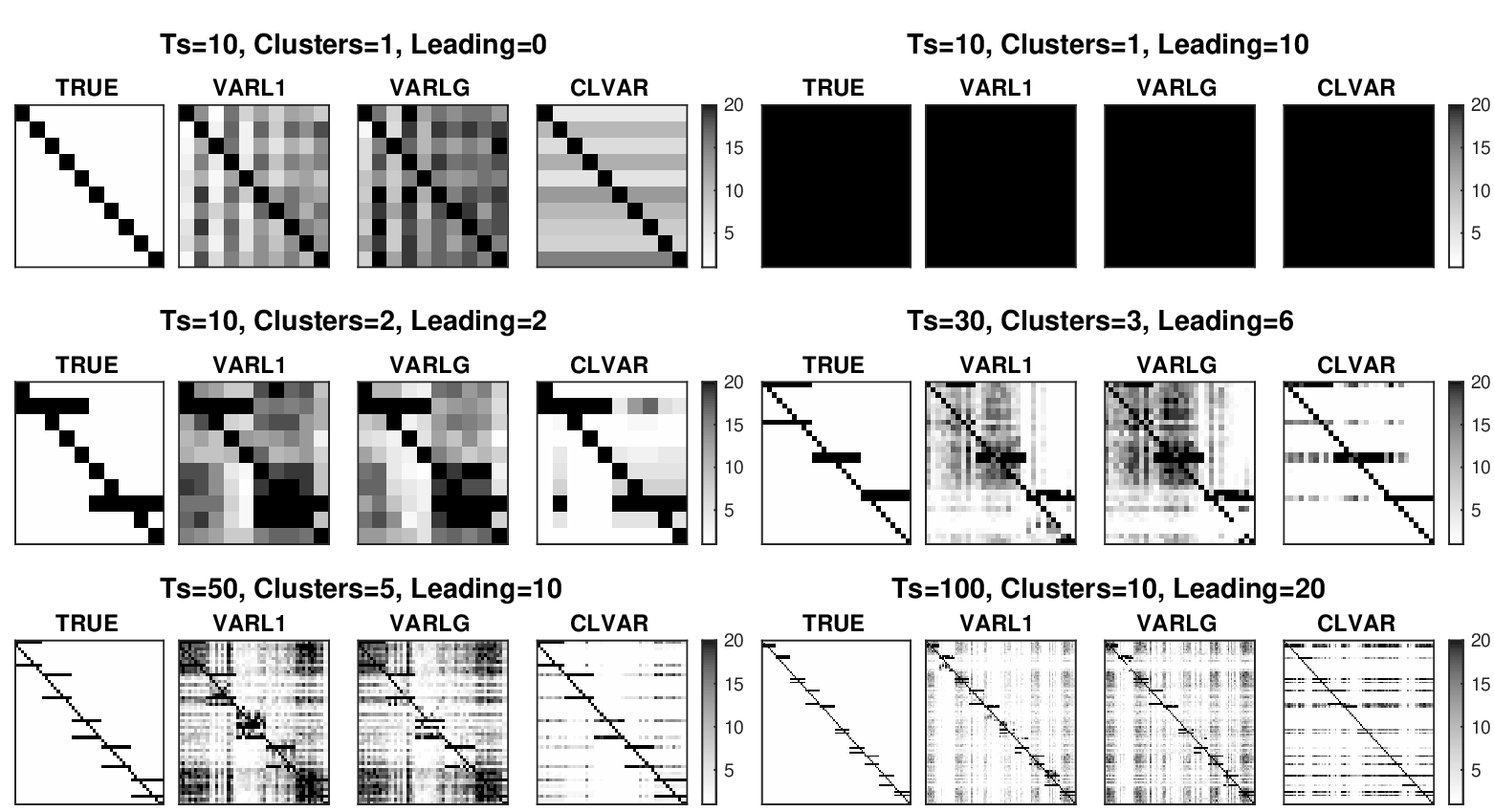}}
\vspace{-5pt}
\caption{Synthesis of model parameters $\vc{W}$}\label{fig:synthGrang}
\vspace{-5pt}
\end{wrapfigure}

To better understand the behaviour of the methods in terms of the structure they learn, we chart in Fig. \ref{fig:synthGrang} a synthesis of the model matrices $\vc{W}$ learned by the sparse learning methods for the largest training size in the 4th system\footnote{For space reasons, results for the other experiments are deferred to the Supplement.}.
The displayed structures correspond to the schema of the $\vc{W}$ matrix presented in Fig.\ref{fig:WMatrix}.
For the figure, the matrices were binarised to simply indicate the existence (1) or non-existence (0) of a G-causal link.
The white-to-black shading reflects the number of experimental replications in which this binary indicator is active (equal to 1).
So, a black element in the matrix means that this G-causal link was learned in all the 20 re-samples of the generating process. White means no G-causality in any of the re-samples.
Though none of the sparse method was able to clearly and systematically recover the true structures, VARL1 and VARLG clearly suffer from more numerous and more frequent over-selections than CLVAR which matches the true structure more closely and with higher selection stability (fewer light-shaded elements).



Finally, we explored how the CLVAR scales with increasing sample size $T$ and the number of time series $K$. The empirical results correspond to the complexity analysis of section \ref{sec:CLVAR}: the run-times increased fairly slowly with increasing sample size $T$ but were much longer for systems with higher number of series $K$.
Further details are deferred to the Supplement.
Overall, the synthetic experiments confirm the desired properties of CLVAR in terms of improved predictive accuracy and structural recovery.

\subsection{Real-data Experiments}
\label{sec:realexperiments}

We used two real datasets very different in nature, frequency and length of available observations.
First, an USGS dataset of daily averages of water physical discharge\footnote{USGS parameter code 00060 - physical discharge in cubic feet per second.} measured at 17 sites along the Yellowstone (8 sites) and Connecticut (9 sites) river streams (source: Water Services of the US geological survey \url{http://www.usgs.gov/}).
Second, an economic dataset of quarterly data on 20 major US macro-economic indicators of \cite{Stock2012} frequently used as a benchmark dataset for VAR learning methods.
More details on the datasets can be found in the Supplement.

We preprocessed the data by standard stationary transformations: we followed \cite{Stock2012} for the economic dataset; by year-on-year log-differences for the USGS.
For the short economic dataset, we fixed the hold-out length to 30 and the training sizes from 50 to 130.
For the much longer USGS dataset, the hold-out is 300 and the training size increases from 200 to 600.
The re-samples are constructed by dropping the latest observation from the data and constructing the shifted train and hold-out from this curtailed dataset.

\begin{figure}[h!]
\centering
\subfigure[MSE and G-causal edges][c]{\label{fig:realResults1}\includegraphics[width=0.42\linewidth]{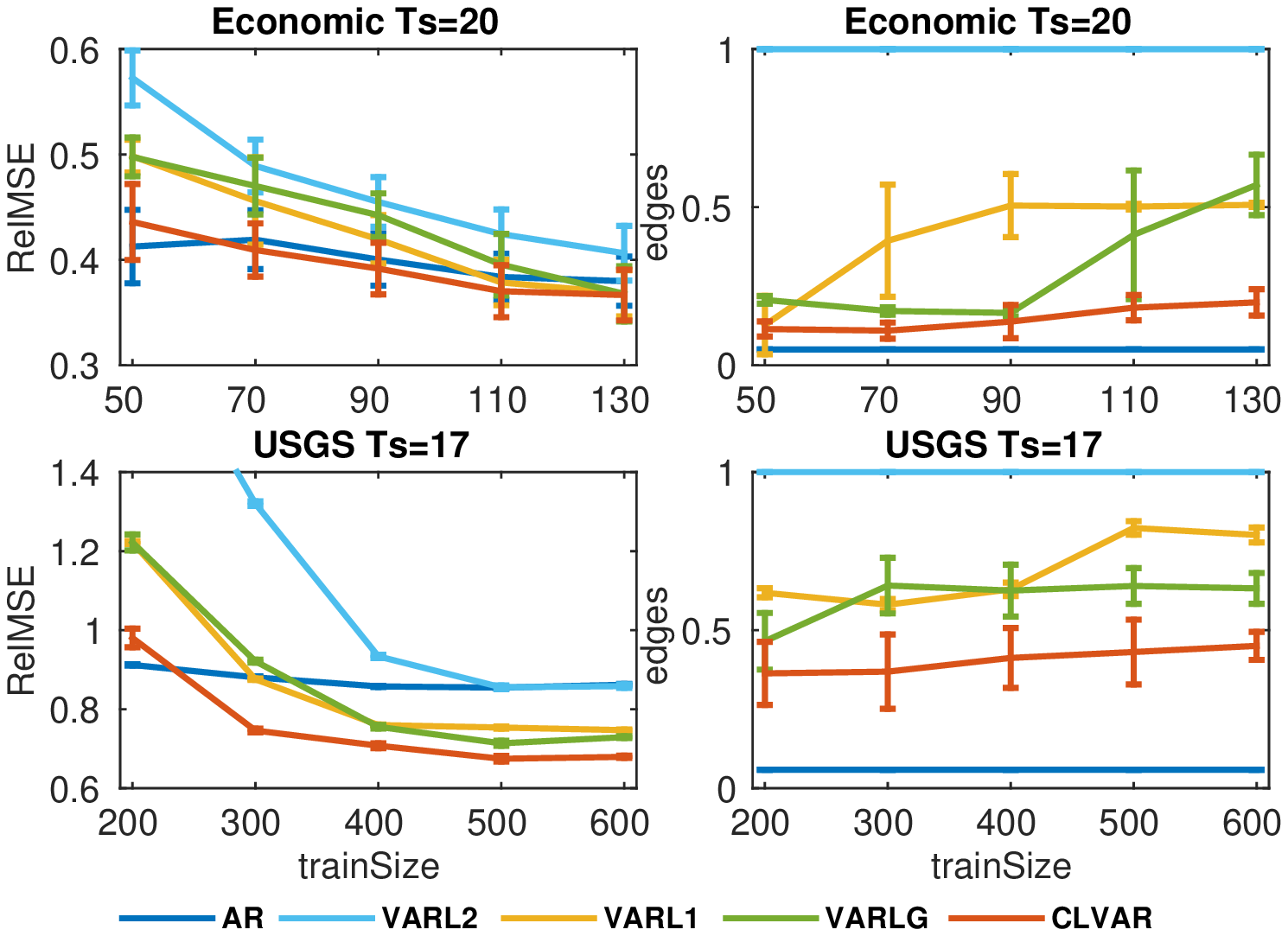}}
\quad \quad
\subfigure[Synthesis of parameters $\vc{W}$][c]{\label{fig:realResults2}\includegraphics[width=0.37\linewidth \vspace{1em}]{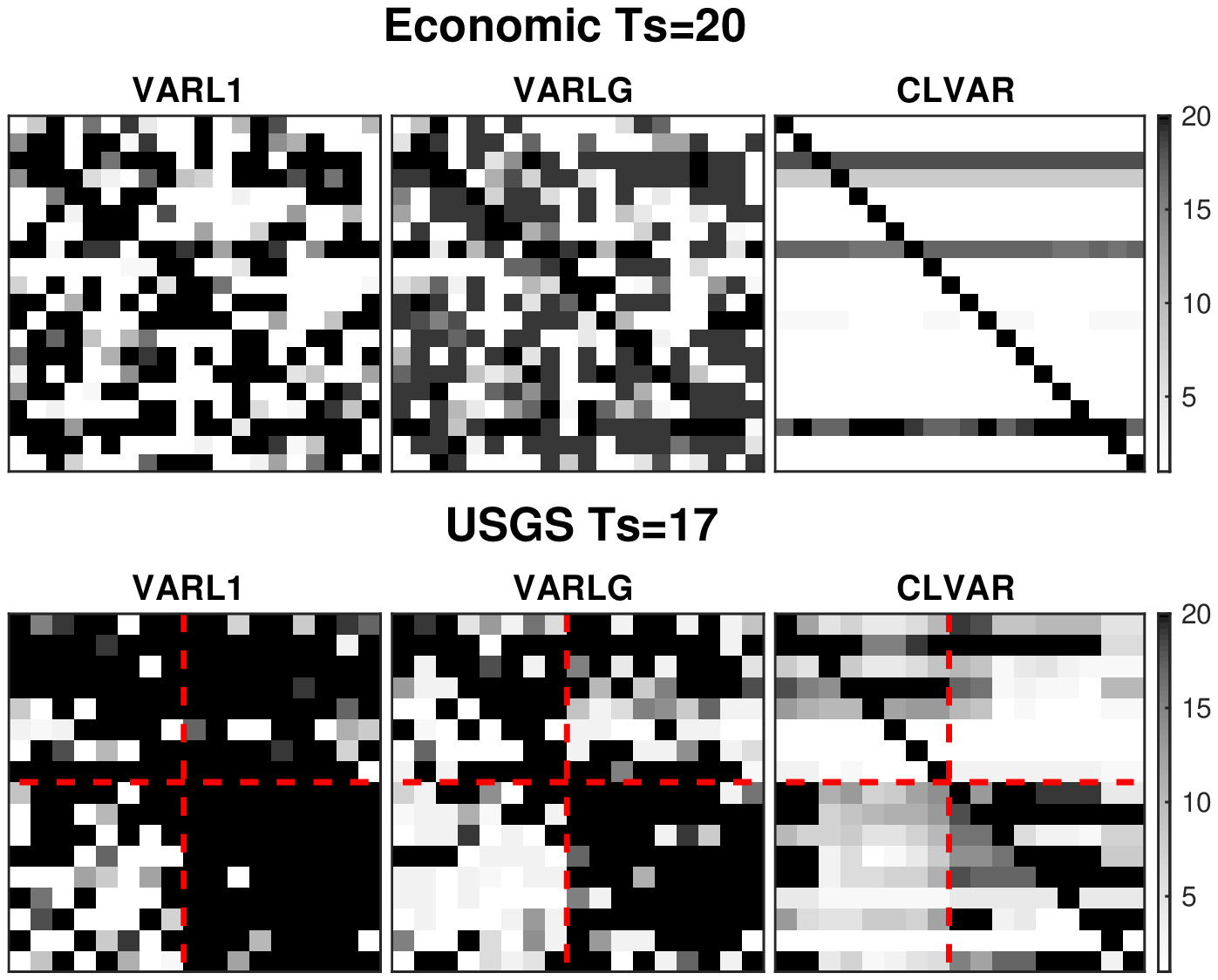}\vspace{1em}}
\caption{Results for real-data experiments averaged over 20 experimental replications} 
\label{fig:realResults}
\vspace{-10pt}
\end{figure}

The results of the two sets of experiments are presented in Fig. \ref{fig:realResults}.
The true parameters of the generative processes are unknown here.
Therefore the predictive accuracy is measured in terms of the MSE relative to a random walk model (the lower the better), and the structural recovery is measured in terms of the proportion of active edges in the G-causal graph (the lower the better), always averaged across the 20 re-samples with $\pm 1$ standard deviation errorbar.

Similarly as in the synthetic experiments, the predictive performance improves with increasing training size and the differences between the methods get smaller.
In both experiments, the non-sparse VARL2 has the worst forecasting accuracy (which corresponds to the initial motivation that real large time-series systems tend to be sparse). 
CLVAR outperformed the other two sparse learning methods VARL1 and VARLG in predictive accuracy as well as sparsity of the learned G-causal graphs. 
In the economic experiment, the completely (by construction) sparse AR achieved similar predictive accuracy. CLVAR clearly outperforms all the other methods on the USGS dataset.

Fig. \ref{fig:realResults2} explores the effect of the structural assumptions on the final shape of the model parameter matrices $\vc{W}$ in the same manner as in Fig. \ref{fig:synthGrang}.
The CLVAR matrices are much sparser than the VARL1 and VARLG matrices, organised around a small number of leading indicators.
In the economic dataset, the CLVAR method identified three leading indicators for the whole system. 
In the USGS dataset, the dashed red lines delimit the the Yellowstone (top-left) from the Connecticut (bottom-right) sites. 
In both these sets of experiments the recovered structure helped improving the forecasts beyond the accuracy achievable by the other tested learning methods.

\section{Conclusions}
\label{sec:conclusions}

We presented here a new method for learning sparse VAR models with shared structures in their Granger causality graphs based on the leading indicators of the system, a problem that had not been previously addressed in the time series literature.

The new method has multiple learning objectives: good forecasting performance of the models, and the discovery of the leading indicators and the clusters of series around them.
Meeting these simultaneously is not trivial and
we used the techniques of multi-task and structured sparsity learning to achieve it.
The method promotes shared patterns in the structure of the individual predictive tasks by forcing them onto a lower-dimensional sub-space spanned by sparse prototypes of the cluster centres.
The empirical evaluation confirmed the efficacy of our approach through favourable results of our new method as compared to the state-of-the-art.



\bibliography{Lead_ACML2017}

%
%
%
%

\appendix

\section{Experimental data and transformations}\label{sec:ExpData}

Table \ref{tab:riversData} lists the measurement sites of the Water Service of the US Geological Survey (\url{http://www.usgs.gov/}) whose data we use in the USGS experiments in section 5.2 of the main text.
The original data are the daily averages of the physical discharge in cubic feet per second (parameter code 00060) downloaded from the USGS database on 9/9/2016.
We have used the data up to 31/12/2014 and before modelling transformed them by taking the year-on-year log-differences.

\begin{small}
\begin{table}[h!]
\caption{Measurement sites for the river-flow data}
\label{tab:riversData}
\vskip 0.1in
\centering
\begin{tabular}{ l | l }
\hline
\hline
Code & Description \\
\hline
\hline
06191500 & Yellowstone River at Corwin Springs MT \\
06192500 & Yellowstone River near Livingston MT \\
06214500 & Yellowstone River at Billings MT \\
06295000 & Yellowstone River at Forsyth MT \\
06309000 & Yellowstone River at Miles City MT \\
06327500 & Yellowstone River at Glendive MT \\
06329500 & Yellowstone River near Sidney MT \\
01129200 & CONNECTICUT R BELOW INDIAN STREAM NR PITTSBURG, NH \\
01129500 & CONNECTICUT RIVER AT NORTH STRATFORD, NH \\
01131500 & CONNECTICUT RIVER NEAR DALTON, NH \\
01138500 & CONNECTICUT RIVER AT WELLS RIVER, VT \\
01144500 & CONNECTICUT RIVER AT WEST LEBANON, NH \\
01154500 & CONNECTICUT RIVER AT NORTH WALPOLE, NH \\
01170500 & CONNECTICUT RIVER AT MONTAGUE CITY, MA \\
01184000 & CONNECTICUT RIVER AT THOMPSONVILLE, CT \\

\hline
\hline
\end{tabular}
\end{table}
\end{small}

Table \ref{tab:SWData} lists the macro-economic indicators of \cite{Stock2012} used in our economic experiment in section 5.2 in the main text.
Before using for modelling we have applied the same pre-processing steps as in \cite{Stock2012}.

We have
\vskip -10pt
\begin{compactitem}
\item transformed the monthly data to quarterly by taking the quarterly averages (column Q in table \ref{tab:SWData});
\item applied the stationarizing transformations described in table \ref{tab:Transform} (column T in table \ref{tab:SWData});
\item cleaned the data from outliers by replacing observations with absolute deviations from median larger than 6 times the interquartile range by the median of the 5 preceding values.
\end{compactitem}

\begin{small}
\begin{table}[h!]
\caption{Stationarizing transformations}
\label{tab:Transform}
\vskip 0.1in
\centering
\begin{tabular}{ l | c }
\hline
\hline
T & Transformation \\
\hline
\hline
1 & $y_t = z_t$\\
2 & $y_t = z_t - z_{t-1}$\\
3 & $y_t = (z_t - z_{t-1}) - (z_{t-1} - z_{t-2})$\\
4 & $y_t = \log(z_t)$\\
5 & $y_t = \ln(z_t/z_{t-1})$\\
6 & $y_t = \ln(z_t/z_{t-1}) - \ln(z_{t-1}/z_{t-2})$\\
\hline
\hline
\multicolumn{2}{l}{\small{$z_t$ is the original data, $y_t$ is the transformed series}}
\end{tabular}
\end{table}
\end{small}

\begin{small}
\begin{table}[h!]
\caption{Macro-economic data and transformations}
\label{tab:SWData}
\vskip 0.1in
\centering
\begin{tabular}{ l | c | c | l}
\hline
\hline
Code & Q & T & Description \\
\hline
\hline
GDP251 & Q & 5 & real gross domestic product, quantity index (2000=100) , saar\\
CPIAUCSL & M & 6 & cpi all items (sa) fred\\
FYFF & M & 2 & interest rate: federal funds (effective) (\%  per annum,nsa)\\
PSCCOMR & M & 5 & real spot mrkt price idx:bls \&  crb: all commod(1967=100)\\
FMRNBA & M & 3 & depository inst reserves:nonborrowed,adj res req chgs(mil\$ ,sa)\\
FMRRA & M & 6 & depository inst reserves:total,adj for reserve req chgs(mil\$ ,sa)\\
FM2 & M & 6 & money stock:m2 (bil\$,sa)\\
GDP252 & Q & 5 & real personal consumpt expend, quantity idx (2000=100) , saar\\
IPS10 & M & 5 & industrial production index -  total index\\
UTL11 & M & 1 & capacity utilization - manufacturing (sic)\\
LHUR & M & 2 & unemployment rate: all workers, 16 years \&  over (\% ,sa)\\
HSFR & M & 4 & housing starts:nonfarm(1947-58),total farm\& nonfarm(1959-)\\
PWFSA & M & 6 & producer price index: finished goods (82=100,sa)\\
GDP273 & Q & 6 & personal consumption expenditures, price idx (2000=100) , saar\\
CES275R & M & 5 & real avg hrly earnings, prod wrkrs, nonfarm - goods-producing\\
FM1 & M & 6 & money stock: m1(bil\$ ,sa)\\
FSPIN & M & 5 & s\& p's common stock price index: industrials (1941-43=10)\\
FYGT10 & M & 2 & interest rate: u.s.treasury const matur,10-yr.(\%  per ann,nsa)\\
EXRUS & M & 5 & united states,effective exchange rate(merm)(index no.)\\
CES002 & M & 5 & employees, nonfarm - total private\\
\hline
\hline
\end{tabular}
\end{table}
\end{small}

\section{Experimental results}\label{sec:ExpResults}

This section provides further details on experimental results not included in the main text due to space limitation.

\subsection{Synthetic experiments}\label{sec:Synthetic}

\begin{figure}[h!]
\centering
\includegraphics[width=1\textwidth]{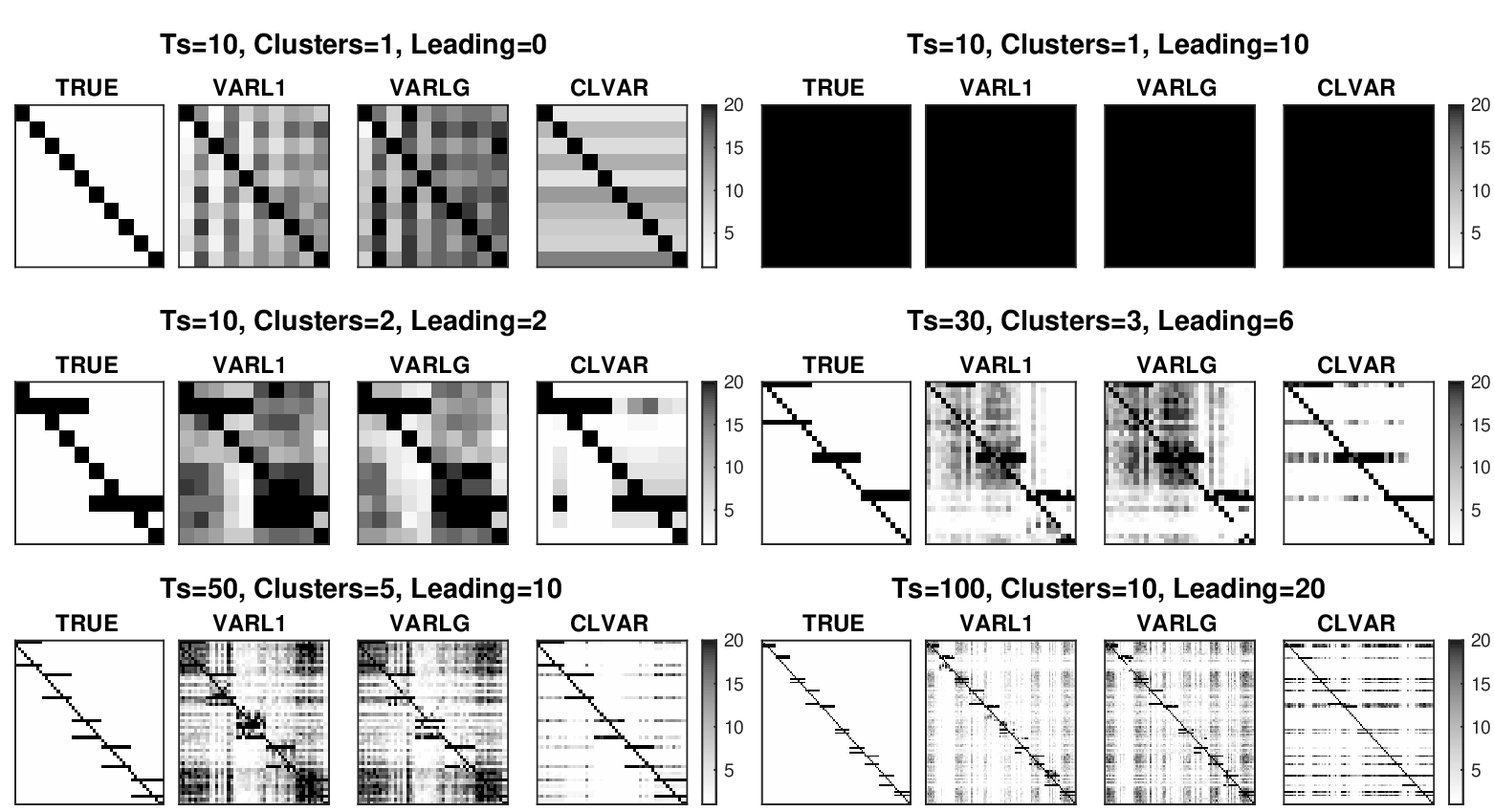}
\caption{Synthesis of model parameters $\vc{W}$}\label{fig:SynthGranger}
\end{figure}

Fig. \ref{fig:SynthGranger} shows the synthesis of the model parameter matrices $\vc{W}$ for the six synthetic experimental designs. The displayed structures correspond to the schema of the $\vc{W}$ matrix presented in Fig. 1 of the main text. For the figure, the matrices were binarised to simply indicate the existence (1) or non-existence (0) of a G-causal link.
The white-to-black shading reflects the number of experimental replications in which this binary indicator is active (equal to 1).
So, a black element in the matrix means that this G-causal link was learned in all the 20 re-samples of the generating process. White means no G-causality in any of the re-samples.
Though none of the sparse method was able to clearly and systematically recover the true structures, VARL1 and VARLG clearly suffer from more numerous and more frequent over-selections than CLVAR which matches the true structures more closely and with higher selection stability (fewer light-shaded elements).
The 4th experimental set-up is included in the main text as Fig. 4.

\begin{figure}[h!]
\centering
\includegraphics[width=1\textwidth]{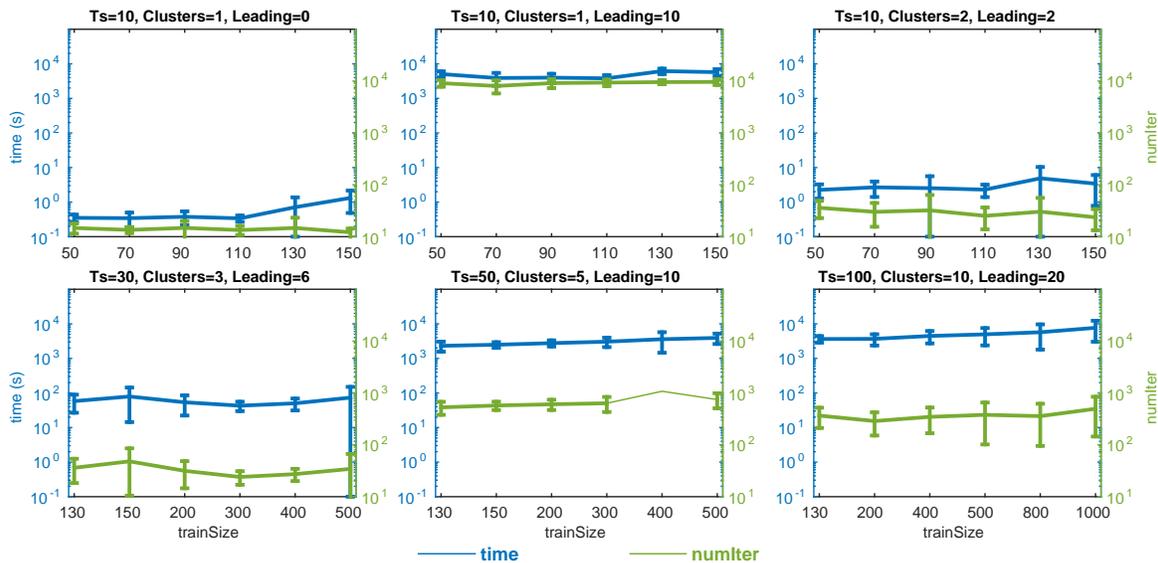}
\caption{Runtime and number of iterations}\label{fig:SynthTime}
\end{figure}

Fig. \ref{fig:SynthTime} summarises the scaling properties of the CLVAR method with increasing increasing sample size $T$ and the number of time series $K$.
In each experiment, we selected a single hyper-parameter combination (near the optimal) and measured the time in seconds (on a single Intel(R) Xeon(R) CPU E5-2680 v2 @ 2.80GHz) and the number of iterations needed till the convergence of the objective (with $10^{-5}$ tolerance) for the 20 data re-samples. 
We used the $\ell_2$ regularised solution as a warm start.
The empirical results correspond to the theoretical complexity analysis of section 3.2 in the main text. For an experimental set-up with fixed number of series $K$ (and G-causal structure), the run-time typically grows fairly slowly with the sample sizes $T$.
However, the increases are much more important when moving to larger experiments, with higher $K$ and more complicated structures.
Here the growth in run-time is accompanied by higher number of iterations.
From our experimental set-up it is difficult to separate the effect of enlarging the time-series
systems in terms of higher $K$ from the effect of more complicated structures in
terms of higher number of clusters and leading indicators.
In reality, we expect these to go hand-in-hand so in this sense our empirical analysis complements the theoretical asymptotic complexity analysis of section 3.2 of the main text

Table \ref{tab:Synth_MSE} provides the numerical data behind the plots of Fig. 3(a) in the main text.
The predictive accuracy is measured by mean squared error (MSE) of 1-step-ahead forecasts relative to the forecasts produced by the VAR with the true generative coefficients (the irreducible error). 
The relative MSE is averaged over the 500 hold-out points (the models are fixed and the forecasts are produced by sliding forward over the dataset).
The \emph{avg} and \emph{std} are the average and standard deviation calculated over the 20 re-samples of the data for each experimental design.

\begin{center}

\begin{longtable}{  l | c | c | c | c | c | c  }
\caption{Synthetic experiments: Relative MSE over true model}\label{tab:Synth_MSE}\\
\hline
trainSize & stat  & AR & VARL2 & VARL1 & VARLG & CLVAR \\
\hline
\endfirsthead

\multicolumn{7}{r}{... continues from previous page} \\
\hline
trainSize & stat  & AR & VARL2 & VARL1 & VARLG & CLVAR \\
\hline
\endhead

\multicolumn{7}{r}{continues in next page ...} \\
\endfoot

\hline \hline
\endlastfoot

\hline \hline 
\multicolumn{7}{c}{Ts=10, Clusters=1, Leading=0}\\
\hline
50 & avg & 1.225 & 3.113 & 1.974 & 2.513 & 1.772\\ 
 & std & 0.102 & 0.637 & 0.310 & 0.515 & 0.643\\ 
70 & avg & 1.143 & 2.570 & 1.503 & 1.868 & 1.395\\ 
 & std & 0.063 & 0.474 & 0.162 & 0.304 & 0.328\\ 
90 & avg & 1.098 & 2.186 & 1.324 & 1.545 & 1.219\\ 
 & std & 0.043 & 0.353 & 0.125 & 0.203 & 0.082\\ 
110 & avg & 1.071 & 1.926 & 1.232 & 1.368 & 1.160\\ 
 & std & 0.030 & 0.268 & 0.086 & 0.133 & 0.055\\ 
130 & avg & 1.052 & 1.745 & 1.172 & 1.267 & 1.121\\ 
 & std & 0.020 & 0.230 & 0.062 & 0.096 & 0.038\\ 
150 & avg & 1.043 & 1.617 & 1.143 & 1.214 & 1.102\\ 
 & std & 0.017 & 0.178 & 0.050 & 0.076 & 0.034\\ 
\hline \hline 
\multicolumn{7}{c}{Ts=10, Clusters=1, Leading=10}\\
\hline
50 & avg & 12.564 & 23.449 & 42.347 & 42.055 & 22.319\\ 
 & std & 5.258 & 18.273 & 32.893 & 32.115 & 15.458\\ 
70 & avg & 12.649 & 12.712 & 9.610 & 9.634 & 11.692\\ 
 & std & 4.507 & 4.565 & 3.681 & 3.472 & 4.769\\ 
90 & avg & 12.281 & 4.905 & 4.943 & 4.897 & 4.934\\ 
 & std & 4.697 & 1.545 & 1.513 & 1.547 & 1.584\\ 
110 & avg & 12.583 & 3.569 & 3.565 & 3.562 & 3.509\\ 
 & std & 4.316 & 0.940 & 0.939 & 0.940 & 0.917\\ 
130 & avg & 12.028 & 2.527 & 2.523 & 2.519 & 2.477\\ 
 & std & 3.762 & 0.480 & 0.479 & 0.477 & 0.486\\ 
150 & avg & 11.637 & 2.195 & 2.194 & 2.189 & 2.158\\ 
 & std & 3.157 & 0.417 & 0.417 & 0.414 & 0.433\\ 
\hline \hline 
\multicolumn{7}{c}{Ts=10, Clusters=2, Leading=2}\\
\hline
50 & avg & 1.429 & 2.730 & 1.743 & 2.124 & 1.674\\ 
 & std & 0.167 & 0.565 & 0.246 & 0.357 & 0.308\\ 
70 & avg & 1.336 & 2.296 & 1.522 & 1.689 & 1.405\\ 
 & std & 0.096 & 0.346 & 0.206 & 0.199 & 0.139\\ 
90 & avg & 1.295 & 1.998 & 1.398 & 1.499 & 1.289\\ 
 & std & 0.100 & 0.253 & 0.139 & 0.146 & 0.116\\ 
110 & avg & 1.258 & 1.824 & 1.320 & 1.396 & 1.199\\ 
 & std & 0.064 & 0.214 & 0.082 & 0.103 & 0.062\\ 
130 & avg & 1.237 & 1.686 & 1.283 & 1.347 & 1.156\\ 
 & std & 0.057 & 0.181 & 0.078 & 0.085 & 0.050\\ 
150 & avg & 1.229 & 1.566 & 1.255 & 1.300 & 1.133\\ 
 & std & 0.054 & 0.171 & 0.076 & 0.081 & 0.043\\ 
\hline \hline 
\multicolumn{7}{c}{Ts=30, Clusters=3, Leading=6}\\
\hline
130 & avg & 1.532 & 2.424 & 1.355 & 1.490 & 1.261\\ 
 & std & 0.142 & 0.361 & 0.091 & 0.131 & 0.081\\ 
150 & avg & 1.512 & 2.245 & 1.319 & 1.414 & 1.225\\ 
 & std & 0.129 & 0.308 & 0.081 & 0.111 & 0.062\\ 
200 & avg & 1.471 & 1.861 & 1.257 & 1.289 & 1.172\\ 
 & std & 0.113 & 0.205 & 0.062 & 0.069 & 0.048\\ 
300 & avg & 1.451 & 1.586 & 1.164 & 1.224 & 1.100\\ 
 & std & 0.103 & 0.135 & 0.039 & 0.051 & 0.031\\ 
400 & avg & 1.438 & 1.375 & 1.119 & 1.151 & 1.073\\ 
 & std & 0.100 & 0.086 & 0.028 & 0.036 & 0.024\\ 
500 & avg & 1.434 & 1.298 & 1.097 & 1.116 & 1.059\\ 
 & std & 0.099 & 0.068 & 0.023 & 0.027 & 0.018\\ 
\hline \hline 
\multicolumn{7}{c}{Ts=50, Clusters=5, Leading=10}\\
\hline
130 & avg & 3.327 & 4.376 & 2.024 & 2.160 & 1.746\\ 
 & std & 0.557 & 0.792 & 0.306 & 0.277 & 0.223\\ 
150 & avg & 3.254 & 3.920 & 1.801 & 1.915 & 1.577\\ 
 & std & 0.544 & 0.685 & 0.205 & 0.215 & 0.196\\ 
200 & avg & 3.200 & 3.203 & 1.583 & 1.620 & 1.324\\ 
 & std & 0.506 & 0.523 & 0.146 & 0.145 & 0.109\\ 
300 & avg & 3.161 & 2.280 & 1.347 & 1.392 & 1.128\\ 
 & std & 0.491 & 0.301 & 0.083 & 0.092 & 0.034\\ 
400 & avg & 3.123 & 1.844 & 1.228 & 1.299 & 1.091\\ 
 & std & 0.478 & 0.192 & 0.053 & 0.070 & 0.025\\ 
500 & avg & 3.097 & 1.622 & 1.172 & 1.214 & 1.081\\ 
 & std & 0.468 & 0.144 & 0.039 & 0.050 & 0.025\\ 
\hline \hline 
\multicolumn{7}{c}{Ts=100, Clusters=10, Leading=20}\\
\hline
130 & avg & 1.890 & 3.362 & 1.518 & 1.807 & 1.734\\ 
 & std & 0.202 & 0.536 & 0.118 & 0.184 & 0.178\\ 
200 & avg & 1.845 & 2.855 & 1.350 & 1.478 & 1.415\\ 
 & std & 0.189 & 0.420 & 0.078 & 0.106 & 0.109\\ 
400 & avg & 1.801 & 2.196 & 1.225 & 1.268 & 1.180\\ 
 & std & 0.177 & 0.264 & 0.050 & 0.059 & 0.049\\ 
600 & avg & 1.783 & 1.820 & 1.137 & 1.191 & 1.108\\ 
 & std & 0.172 & 0.182 & 0.031 & 0.043 & 0.026\\ 
800 & avg & 1.777 & 1.644 & 1.104 & 1.129 & 1.084\\ 
 & std & 0.170 & 0.142 & 0.023 & 0.029 & 0.021\\ 
1000 & avg & 1.774 & 1.501 & 1.090 & 1.100 & 1.065\\ 
 & std & 0.170 & 0.115 & 0.020 & 0.022 & 0.019\\

\end{longtable}
\end{center}

Table \ref{tab:Synth_Spars} provides the numerical data behind the plots of Fig. 3(b) in the main text.
The selection accuracy of the true G-causal links is measured by the average between the false negative and false positive rates.
The \emph{avg} and \emph{std} are the average and standard deviation calculated over the 20 re-samples of the data for each experimental design.

\begin{center}

\begin{longtable}{  l | c | c | c | c | c | c  }
\caption{Synthetic experiments: Selection errors of true G-causal links}\label{tab:Synth_Spars}\\
\hline
trainSize & stat  & AR & VARL2 & VARL1 & VARLG & CLVAR \\
\hline
\endfirsthead

\multicolumn{7}{r}{... continues from previous page} \\
\hline
trainSize & stat  & AR & VARL2 & VARL1 & VARLG & CLVAR \\
\hline
\endhead

\multicolumn{7}{r}{continues in next page ...} \\
\endfoot

\hline \hline
\endlastfoot

\hline \hline 
\multicolumn{7}{c}{Ts=10, Clusters=1, Leading=0}\\
\hline
50 & avg & 0.000 & 0.500 & 0.268 & 0.316 & 0.151\\ 
 & std & 0.000 & 0.000 & 0.062 & 0.072 & 0.153\\ 
70 & avg & 0.000 & 0.500 & 0.245 & 0.318 & 0.119\\ 
 & std & 0.000 & 0.000 & 0.027 & 0.056 & 0.094\\ 
90 & avg & 0.000 & 0.500 & 0.207 & 0.301 & 0.118\\ 
 & std & 0.000 & 0.000 & 0.027 & 0.026 & 0.043\\ 
110 & avg & 0.000 & 0.500 & 0.172 & 0.285 & 0.122\\ 
 & std & 0.000 & 0.000 & 0.026 & 0.027 & 0.034\\ 
130 & avg & 0.000 & 0.500 & 0.141 & 0.268 & 0.138\\ 
 & std & 0.000 & 0.000 & 0.017 & 0.029 & 0.036\\ 
150 & avg & 0.000 & 0.500 & 0.120 & 0.245 & 0.143\\ 
 & std & 0.000 & 0.000 & 0.019 & 0.027 & 0.041\\ 
\hline \hline 
\multicolumn{7}{c}{Ts=10, Clusters=1, Leading=10}\\
\hline
50 & avg & 0.450 & 0.000 & 0.027 & 0.031 & 0.043\\ 
 & std & 0.000 & 0.000 & 0.084 & 0.095 & 0.112\\ 
70 & avg & 0.450 & 0.000 & 0.013 & 0.013 & 0.010\\ 
 & std & 0.000 & 0.000 & 0.054 & 0.058 & 0.026\\ 
90 & avg & 0.450 & 0.000 & 0.001 & 0.000 & 0.017\\ 
 & std & 0.000 & 0.000 & 0.005 & 0.000 & 0.042\\ 
110 & avg & 0.450 & 0.000 & 0.000 & 0.000 & 0.000\\ 
 & std & 0.000 & 0.000 & 0.000 & 0.000 & 0.000\\ 
130 & avg & 0.450 & 0.000 & 0.000 & 0.000 & 0.000\\ 
 & std & 0.000 & 0.000 & 0.000 & 0.000 & 0.000\\ 
150 & avg & 0.450 & 0.000 & 0.000 & 0.000 & 0.000\\ 
 & std & 0.000 & 0.000 & 0.000 & 0.000 & 0.000\\ 
\hline \hline 
\multicolumn{7}{c}{Ts=10, Clusters=2, Leading=2}\\
\hline
50 & avg & 0.222 & 0.500 & 0.190 & 0.268 & 0.168\\ 
 & std & 0.000 & 0.000 & 0.041 & 0.041 & 0.103\\ 
70 & avg & 0.222 & 0.500 & 0.179 & 0.248 & 0.096\\ 
 & std & 0.000 & 0.000 & 0.072 & 0.039 & 0.061\\ 
90 & avg & 0.222 & 0.500 & 0.159 & 0.222 & 0.094\\ 
 & std & 0.000 & 0.000 & 0.067 & 0.030 & 0.051\\ 
110 & avg & 0.222 & 0.500 & 0.155 & 0.200 & 0.058\\ 
 & std & 0.000 & 0.000 & 0.086 & 0.031 & 0.044\\ 
130 & avg & 0.222 & 0.500 & 0.192 & 0.205 & 0.067\\ 
 & std & 0.000 & 0.000 & 0.107 & 0.076 & 0.054\\ 
150 & avg & 0.222 & 0.500 & 0.210 & 0.196 & 0.054\\ 
 & std & 0.000 & 0.000 & 0.118 & 0.064 & 0.034\\ 
\hline \hline 
\multicolumn{7}{c}{Ts=30, Clusters=3, Leading=6}\\
\hline
130 & avg & 0.321 & 0.500 & 0.175 & 0.204 & 0.173\\ 
 & std & 0.000 & 0.000 & 0.015 & 0.017 & 0.037\\ 
150 & avg & 0.321 & 0.500 & 0.165 & 0.192 & 0.176\\ 
 & std & 0.000 & 0.000 & 0.016 & 0.018 & 0.027\\ 
200 & avg & 0.321 & 0.500 & 0.178 & 0.183 & 0.165\\ 
 & std & 0.000 & 0.000 & 0.046 & 0.014 & 0.029\\ 
300 & avg & 0.321 & 0.500 & 0.225 & 0.166 & 0.107\\ 
 & std & 0.000 & 0.000 & 0.012 & 0.027 & 0.034\\ 
400 & avg & 0.321 & 0.500 & 0.202 & 0.242 & 0.098\\ 
 & std & 0.000 & 0.000 & 0.013 & 0.012 & 0.029\\ 
500 & avg & 0.321 & 0.500 & 0.184 & 0.220 & 0.095\\ 
 & std & 0.000 & 0.000 & 0.009 & 0.013 & 0.022\\ 
\hline \hline 
\multicolumn{7}{c}{Ts=50, Clusters=5, Leading=10}\\
\hline
130 & avg & 0.321 & 0.500 & 0.226 & 0.226 & 0.123\\ 
 & std & 0.000 & 0.000 & 0.040 & 0.010 & 0.032\\ 
150 & avg & 0.321 & 0.500 & 0.211 & 0.217 & 0.110\\ 
 & std & 0.000 & 0.000 & 0.039 & 0.011 & 0.035\\ 
200 & avg & 0.321 & 0.500 & 0.212 & 0.196 & 0.077\\ 
 & std & 0.000 & 0.000 & 0.047 & 0.009 & 0.027\\ 
300 & avg & 0.321 & 0.500 & 0.257 & 0.169 & 0.038\\ 
 & std & 0.000 & 0.000 & 0.011 & 0.021 & 0.008\\ 
400 & avg & 0.321 & 0.500 & 0.236 & 0.183 & 0.034\\ 
 & std & 0.000 & 0.000 & 0.012 & 0.047 & 0.011\\ 
500 & avg & 0.321 & 0.500 & 0.224 & 0.218 & 0.033\\ 
 & std & 0.000 & 0.000 & 0.011 & 0.033 & 0.011\\ 
\hline \hline 
\multicolumn{7}{c}{Ts=100, Clusters=10, Leading=20}\\
\hline
130 & avg & 0.321 & 0.500 & 0.120 & 0.164 & 0.260\\ 
 & std & 0.000 & 0.000 & 0.006 & 0.009 & 0.054\\ 
200 & avg & 0.321 & 0.500 & 0.098 & 0.132 & 0.159\\ 
 & std & 0.000 & 0.000 & 0.006 & 0.006 & 0.028\\ 
400 & avg & 0.321 & 0.500 & 0.158 & 0.094 & 0.121\\ 
 & std & 0.000 & 0.000 & 0.006 & 0.006 & 0.016\\ 
600 & avg & 0.321 & 0.500 & 0.129 & 0.175 & 0.104\\ 
 & std & 0.000 & 0.000 & 0.004 & 0.005 & 0.011\\ 
800 & avg & 0.321 & 0.500 & 0.110 & 0.154 & 0.100\\ 
 & std & 0.000 & 0.000 & 0.004 & 0.005 & 0.011\\ 
1000 & avg & 0.321 & 0.500 & 0.096 & 0.137 & 0.090\\ 
 & std & 0.000 & 0.000 & 0.004 & 0.004 & 0.013

\end{longtable}
\end{center}

\subsection{Real-data experiments}\label{sec:Real}

Table \ref{tab:Real_MSE} provides the numerical data behind the plots of Fig. 5(a) in the main text.
The predictive accuracy is measured by mean squared error (MSE) of 1-step-ahead forecasts relative to the forecasts produced random walk model (= uses the last observed value as the 1-step-ahead forecast). 
The relative MSE is averaged over the 30 and 300 hold-out points for the Economic and the USGS dataset respectively (the models are fixed and the forecasts are produced by sliding forward over the dataset).
The \emph{avg} and \emph{std} are the average and standard deviation calculated over the 20 re-samples of the data for each experimental dataset.

\begin{center}

\begin{longtable}{  l | c | c | c | c | c | c  }
\caption{Real-data experiments: Relative MSE over random walk}\label{tab:Real_MSE}\\
\hline
trainSize & stat  & AR & VARL2 & VARL1 & VARLG & CLVAR \\
\hline
\endfirsthead

\multicolumn{7}{r}{... continues from previous page} \\
\hline
trainSize & stat  & AR & VARL2 & VARL1 & VARLG & CLVAR \\
\hline
\endhead

\multicolumn{7}{r}{continues in next page ...} \\
\endfoot

\hline \hline
\endlastfoot

\hline \hline 
\multicolumn{7}{c}{Economic Ts=20}\\
\hline 
50 & avg & 0.413 & 0.573 & 0.498 & 0.498 & 0.436\\ 
 & std & 0.035 & 0.026 & 0.016 & 0.018 & 0.036\\ 
70 & avg & 0.419 & 0.489 & 0.456 & 0.470 & 0.409\\ 
 & std & 0.028 & 0.025 & 0.041 & 0.027 & 0.025\\ 
90 & avg & 0.400 & 0.455 & 0.420 & 0.442 & 0.392\\ 
 & std & 0.025 & 0.024 & 0.023 & 0.021 & 0.025\\ 
110 & avg & 0.384 & 0.424 & 0.379 & 0.395 & 0.370\\ 
 & std & 0.022 & 0.024 & 0.022 & 0.029 & 0.025\\ 
130 & avg & 0.380 & 0.406 & 0.368 & 0.368 & 0.367\\ 
 & std & 0.023 & 0.026 & 0.022 & 0.026 & 0.024\\ 
\hline \hline 
\multicolumn{7}{c}{USGS Ts=17}\\
\hline 
200 & avg & 0.912 & 1.857 & 1.222 & 1.222 & 0.980\\ 
 & std & 0.013 & 0.098 & 0.058 & 0.046 & 0.069\\ 
300 & avg & 0.881 & 1.320 & 0.876 & 0.922 & 0.746\\ 
 & std & 0.003 & 0.040 & 0.012 & 0.019 & 0.035\\ 
400 & avg & 0.858 & 0.934 & 0.760 & 0.756 & 0.708\\ 
 & std & 0.001 & 0.030 & 0.026 & 0.012 & 0.019\\ 
500 & avg & 0.855 & 0.855 & 0.754 & 0.714 & 0.675\\ 
 & std & 0.003 & 0.021 & 0.011 & 0.007 & 0.015\\ 
600 & avg & 0.862 & 0.858 & 0.747 & 0.729 & 0.679\\ 
 & std & 0.002 & 0.004 & 0.004 & 0.020 & 0.024\\

\end{longtable}
\end{center}

Table \ref{tab:Real_Spars} provides the numerical data behind the plots of Fig. 5(b) in the main text.
The sparsity of the learned models is measured by the proportion of active edges in the learned G-causality graph. 
The \emph{avg} and \emph{std} are the average and standard deviation calculated over the 20 re-samples of the data for each experimental dataset.

\begin{center}

\begin{longtable}{  l | c | c | c | c | c | c  }
\caption{Real-data experiments: proportion of G-causal graph edges}\label{tab:Real_Spars}\\
\hline
trainSize & stat  & AR & VARL2 & VARL1 & VARLG & CLVAR \\
\hline
\endfirsthead

\multicolumn{7}{r}{... continues from previous page} \\
\hline
trainSize & stat  & AR & VARL2 & VARL1 & VARLG & CLVAR \\
\hline
\endhead

\multicolumn{7}{r}{continues in next page ...} \\
\endfoot

\hline \hline
\endlastfoot

\hline \hline
\multicolumn{7}{c}{Economic Ts=20}\\ 
\hline  
50 & avg  &0.050 &1.000 &0.127 &0.207 & 0.115\\ 
  & std  &0.000 &0.000 &0.092 &0.012 & 0.024\\ 
70 & avg  &0.050 &1.000 &0.394 &0.172 & 0.110\\ 
  & std  &0.000 &0.000 &0.177 &0.012 & 0.026\\ 
90 & avg  &0.050 &1.000 &0.505 &0.166 & 0.138\\ 
  & std  &0.000 &0.000 &0.100 &0.011 & 0.053\\ 
110 & avg  &0.050 &1.000 &0.502 &0.412 & 0.182\\ 
  & std  &0.000 &0.000 &0.009 &0.204 & 0.040\\ 
130 & avg  &0.050 &1.000 &0.507 &0.570 & 0.199\\ 
  & std  &0.000 &0.000 &0.007 &0.096 & 0.042\\ 
\hline \hline  
\multicolumn{7}{c}{USGS Ts=17}\\ 
\hline  
200 & avg  &0.059 &1.000 &0.618 &0.465 & 0.363\\ 
  & std  &0.000 &0.000 &0.117 &0.061 & 0.057\\ 
300 & avg  &0.059 &1.000 &0.581 &0.641 & 0.369\\ 
  & std  &0.000 &0.000 &0.018 &0.010 & 0.081\\ 
400 & avg  &0.059 &1.000 &0.629 &0.625 & 0.412\\ 
  & std  &0.000 &0.000 &0.143 &0.005 & 0.060\\ 
500 & avg  &0.059 &1.000 &0.823 &0.640 & 0.431\\ 
  & std  &0.000 &0.000 &0.061 &0.009 & 0.057\\ 
600 & avg  &0.059 &1.000 &0.801 &0.632 & 0.450\\ 
  & std  &0.000 &0.000 &0.014 &0.090 & 0.020\\

\end{longtable}
\end{center}

\end{document}